%% file: main.tex
\documentclass[10pt,twocolumn,letterpaper]{article}

\usepackage[pagenumbers]{iccv} 


\input{config}

\usepackage{url}
\usepackage{tcolorbox}
\tcbuselibrary{breakable}
\usepackage[misc]{ifsym}
\usepackage{fontawesome5}
\usepackage{enumitem}
\setlist[itemize]{label=$\bullet$}
\usepackage{float}

\definecolor{iccvblue}{rgb}{0.21,0.49,0.74}
\definecolor{brightyellow}{RGB}{255, 255, 84}
\definecolor{brightred}{RGB}{234, 51, 35}
\usepackage[pagebackref,breaklinks,colorlinks,allcolors=iccvblue]{hyperref}

\def\paperID{12912} 
\def\confName{ICCV}
\def\confYear{2025}

\title{LongViTU: Instruction Tuning for Long-Form Video Understanding}

\author{Rujie Wu$^{1}$,~~Xiaojian Ma$^{2\dagger}$,~~Hai Ci$^{3}$,~~Yue Fan$^{2}$, \\
Yuxuan Wang$^{2}$,~~Haozhe Zhao$^{1}$,~~Qing Li$^{2}$,~~Yizhou Wang$^{1}$ \\
$^{1}$School of Computer Science, Peking University \\
$^{2}$National Key Laboratory of General Artificial Intelligence, BIGAI \\
$^{3}$National University of Singapore \\
$^{\dagger}$Research lead~\quad~\faGithub~\href{https://rujiewu.github.io/LongViTU.github.io/}{Project page}
}

\begin{document}

\twocolumn[{%
\renewcommand\twocolumn[1][]{#1}%
\maketitle
\begin{center}
    \centering
    \captionsetup{type=figure}
    \includegraphics[width=1\textwidth]{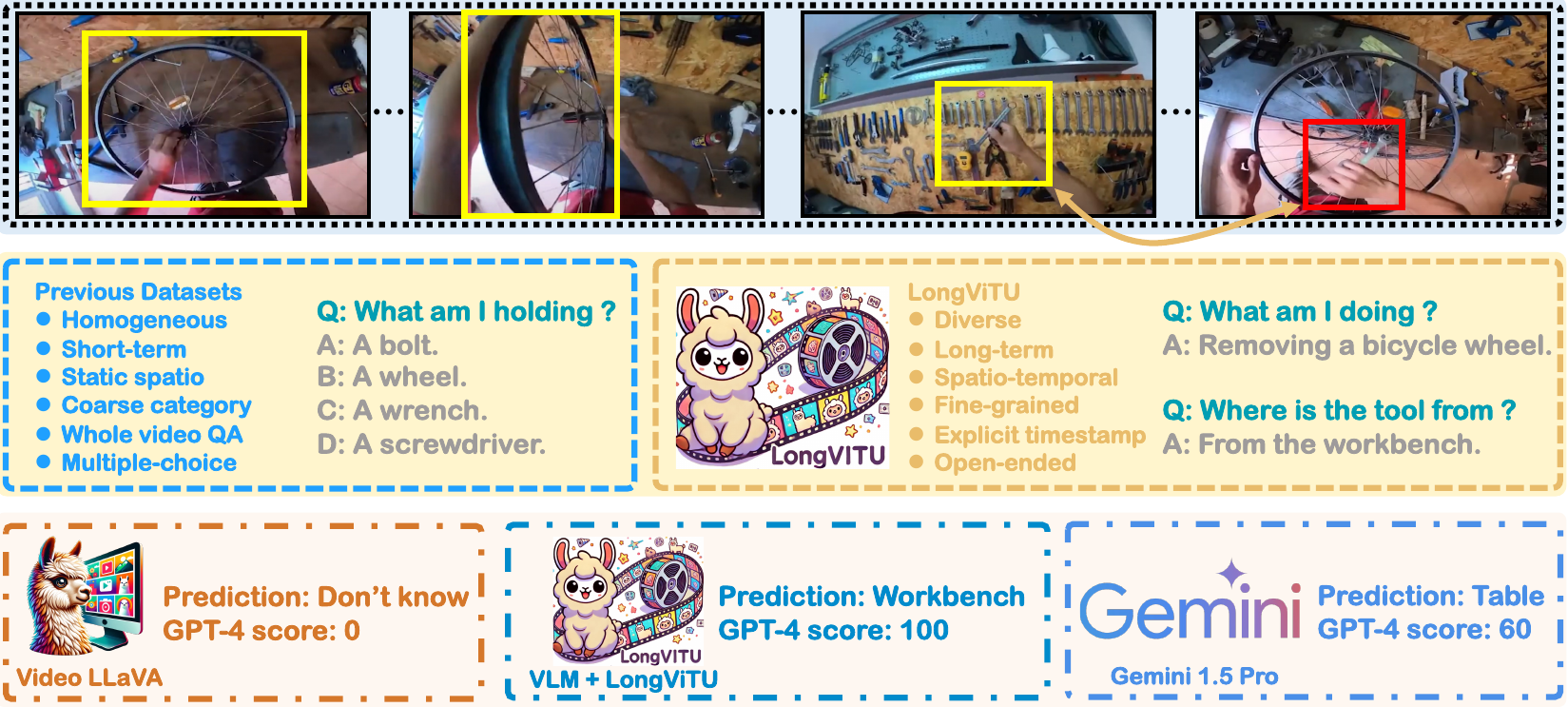}
    \caption{\textbf{Illustration of LongViTU.}~~The top row shows an example video sequence, with \colorbox{brightred!60}{red boxes} highlighting key clues for the posed question and \colorbox{brightyellow!60}{yellow boxes} marking objects in key frames related to the answer. The middle row emphasizes the primary advantages of our proposed LongViTU over previous datasets, along with QA examples; refer to~\autoref{sec:introduction} for further details. The bottom row displays predictions from canonical open-source and proprietary VLMs, evaluated by GPT-4 against ground truth using our novel predefined criteria.}
    \label{fig:teaser}
\end{center}%
}]

\begin{abstract}
This paper introduces LongViTU, a large-scale (\textasciitilde121k QA pairs, \textasciitilde900h videos), automatically generated dataset for long-form video understanding. We propose a \textit{systematic} approach that organizes videos into a \textbf{\textit{hierarchical tree}} structure for QA generation and incorporates \textbf{\textit{self-revision}} mechanisms to ensure high-quality QA pairs. Each QA pair in LongViTU features: 1) long-term context (average \textit{certificate length} of 4.6 minutes); 2) rich knowledge and condensed reasoning (commonsense, causality, planning, \etc). We also offer explicit timestamp annotations of relevant events for each QA pair. We have conducted extensive human studies on LongViTU, and the results prove the quality of our dataset. 
To better evaluate the challenges posed by LongViTU’s emphasis on long-term context and condensed reasoning, we manually curate a subset of LongViTU into a benchmark. Evaluations using a state-of-the-art open-source model (LongVU), a proprietary model (Gemini-1.5-Pro), and human annotators yield GPT-4 scores of 49.9, 52.3, and 81.0, respectively, underscoring the substantial difficulty presented by LongViTU questions. Performing supervised fine-tuning (SFT) of LongVU and LLaVA-Video on LongViTU data results in average performance gains of 2.5\% and 3.7\%, respectively, across a suite of long video understanding benchmarks (EgoSchema, VideoMME-Long, MLVU, LVBench).
\end{abstract}

\begin{table*}[t!]
    \centering
    \caption{\textbf{Comparison with Existing Video QA Datasets.}~~The video sources for each dataset are listed under "Video Source", where "Collection" indicates that videos are derived from various sources. Furthermore,~MC$^{*}$~denotes multiple-choice answers, while~OE$^{**}$~indicates open-ended answers, LongViTU is the first large-scale dataset designed for long-form video understanding with explicit timestamp annotations. $^{\dagger}$The average certificate length, video durations and the number of QA pairs are approximate.}
    \centering
    \resizebox{1\linewidth}{!}{%
    \begin{tabular}{l|c|c|c|c|c|c|c|c}
    \toprule
    \multirow{2}{*}{\textbf{Dataset}} & \multirow{2}{*}{\textbf{Video Source}} & \multirow{2}{*}{\makecell{\textbf{Scenario}}} & \multirow{2}{*}{\textbf{\makecell{Answer\\Types}}} & \multirow{2}{*}{\textbf{\makecell{Condensed\\Reasoning}}} & \multirow{2}{*}{\textbf{\makecell{Explicit\\Timestamp}}} & \multirow{2}{*}{\textbf{\makecell{Certificate \\Length$^{\dagger}$}}} & \multirow{2}{*}{\textbf{\makecell{Video\\Duration}}} & \multirow{2}{*}{\textbf{QAs}} \\
    & & & & & & & \\
    \midrule
    
    EgoVQA~\citep{egovqa} & IU Multiview~\citep{iumultiview} & real world & ~~MC$^{*}$ & \xmark & \xmark & 10s & 10h & 600 \\
    Env-QA~\citep{envqa} & AI2-THOR~\citep{ai2thor} & virtual env & ~~~OE$^{**}$ & \xmark & \xmark & 10s & 130h & 85.1K \\
    EgoTaskQA~\citep{egotaskqa} & LEMMA~\citep{lemma} & real world & OE & \cmark & \xmark & 10s & 15h & 40K \\
    \midrule
    
    MoVQA~\citep{movqa} & Collection & movie & MC & \cmark & \xmark & 230s & 50h & 22K \\
    MovieChat~\citep{moviechat} & Collection & movie & OE & \xmark & \xmark & 60s & 160h & 13K \\
    CinePile~\citep{cinepile} & Collection & movie & MC & \xmark & \xmark & 60s & 420h & 303K \\
    \midrule
    
    \multirow{2}{*}{OpenEQA~\citep{openeqa}} & \multirow{2}{*}{\makecell{ScanNet~\citep{scannet}\\HM3D~\citep{hm3d}}} & \multirow{2}{*}{virtual env} & \multirow{2}{*}{OE} & \multirow{2}{*}{\xmark} & \multirow{2}{*}{\xmark} & \multirow{2}{*}{30s} & \multirow{2}{*}{3h} & \multirow{2}{*}{1.6K} \\
    & & & & & & & \\
    \midrule
    
    WorldQA~\citep{worldqa} & PVSG~\citep{pvsg} & real world & OE & \xmark & \xmark & 60s & 10h & 1K \\
    EgoSchema~\citep{egoschema} & Ego4D~\citep{ego4d} & real world & ~MC & \xmark & \xmark & 100s & 250h & 5K \\
    LLaVA-Video~\citep{llava-video} & Collection & real world & OE & \cmark & \xmark & 30s & 2000h & 960K \\
    MM-Ego~\citep{mmego} & Ego4D~\citep{ego4d} & real world & OE & \xmark & \xmark & 10s & 3000h & 7M \\
    \midrule
    
    \cellcolor{blue!20}\textbf{LongViTU (ours)} & \cellcolor{blue!20}Ego4D~\citep{ego4d} & \cellcolor{blue!20}real world & \cellcolor{blue!20}OE & \cellcolor{blue!20}\cmark & \cellcolor{blue!20}\cmark & \cellcolor{blue!20}276.8s & \cellcolor{blue!20}900h & \cellcolor{blue!20}121K \\
    \bottomrule
    \end{tabular}
    }
    \label{tab:comparison}
\end{table*}%

\section{Introduction}
\label{sec:introduction}

We present LongViTU, a new dataset for large-scale, long-form video understanding (see~\autoref{fig:teaser}) that can be generated fully autonomously using LLM. Compared to existing \textit{Video Question-Answering (Video QA)} datasets, LongViTU emphasizes \textbf{long video context} and \textbf{condensed reasoning}. The primary advantages are summarized below, further details can be found in~\autoref{tab:comparison}.

\setlength{\leftmargini}{0.85em}
\begin{itemize}[topsep=0pt, itemsep=0pt]
\item \textbf{Long Certificate Length.}~~Many Video QA datasets such as NextQA~\citep{nextqa} and ActivityNet-QA~\citep{activitynetqa} have a relatively short average \textit{certificate length} (introduced by EgoSchema~\citep{egoschema} to characterize the length of the video context required to answer the question), typically less than 10 seconds. While datasets with longer videos like WorldQA and EgoSchema~\citep{egoschema}~\citep{worldqa} can reach an average certificate length of 60\textasciitilde100 seconds, LongViTU attains an average certificate length of 276.8 seconds (\textasciitilde4.6 minutes) thanks to our proposed hierarchical tree-structured representations of videos in the data construction pipeline, highlighting its focus on long-context video understanding. For further statistical details, please refer to~\autoref{fig:statistics}.



\item \textbf{Condensed Reasoning.}~~Existing QA datasets~\citep{egotaskqa,openeqa} primarily focus on fundamental spatial elements (e.g., objects, attributes, locations, and states) but do not sufficiently emphasize reasoning. To facilitate questions that require condensed reasoning, we introduce a structured reasoning taxonomy that explicitly guides question generation by categorizing each question according to the specific type of reasoning it demands (e.g., affordance, planning, risk, functionality, causality). By prompting the LLM to align questions with these targeted reasoning categories, we significantly enhance the dataset’s capacity to assess condensed and complex reasoning capabilities.

\item \textbf{Explicit Timestamp Annotations.}~~Previous datasets, such as Otter~\citep{otter}, Video-ChatGPT~\citep{videochatgpt}, InternVideo~\citep{internvideo}, VideoChat~\citep{videochat}, MVBench~\citep{mvbench}, LLaVA-Video~\citep{llava-video}, and MM-Ego~\citep{mmego}, lack explicit timestamp annotations for QA-related events. This omission means the precise start and end times for the relevant video context of each QA are undefined, despite multiple QAs being present in a single video. To address this, our tree-structured video representation enables QA generation at various granularities with explicit timestamps for each event. LongViTU provides precise temporal annotations for all QA events, providing this useful information for identifying key moments in ultra-long video understanding.

\end{itemize}

\noindent
The final LongViTU dataset comprises \textasciitilde121k high-quality QA pairs within \textasciitilde900 hours of videos. To the best of our knowledge, LongViTU is the first publicly available, automatically generated long-form video question-answering dataset featuring condensed reasoning and explicit QA-related timestamp annotations. Our main contributions are summarized as follows:

\setlength{\leftmargini}{0.85em}
\begin{itemize}[topsep=0pt, itemsep=0pt]
    \item We propose a novel automatic pipeline for generating high-quality video question-answering data, with a emphasis on long video context and condensed reasoning.
    \item With our pipeline, we curate LongViTU, a large-scale, high-quality pre-training dataset, accompanied by a manually verified benchmark subset aimed at advancing long-form video understanding.
    \item We conducted extensive experiments on a broad range of long video understanding benchmarks, demonstrating substantial benefits of LongViTU for canonical open-source VLMs, and providing insightful analyses.
\end{itemize}

\begin{figure*}[htbp]
\centering
\includegraphics[width=1\textwidth]{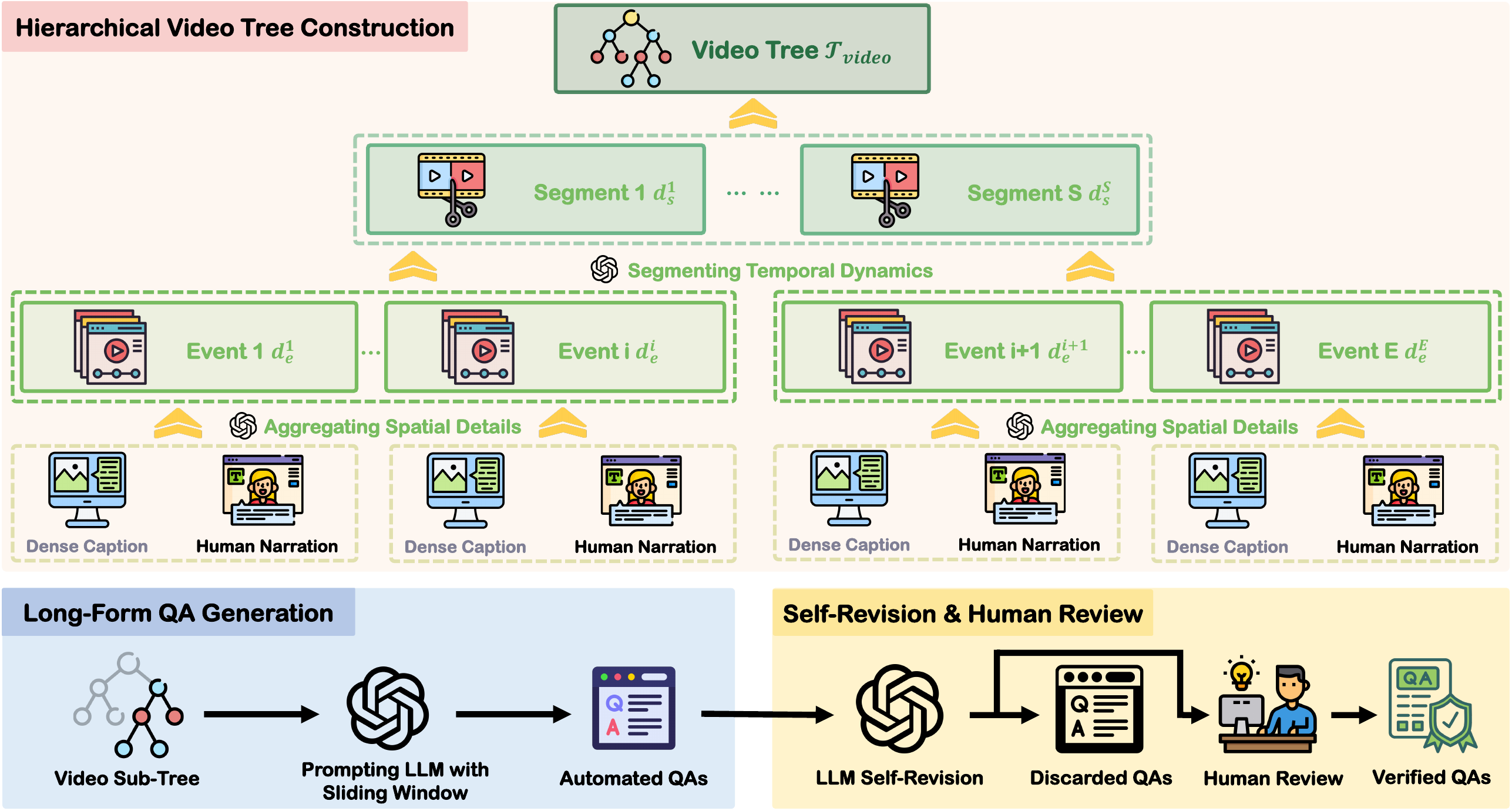}
\caption{\textbf{The construction of LongViTU.}~~We design an automatic pipeline to generate QA pairs from Ego4D~\cite{ego4d} videos while addressing the challenge of long-form video comprehension. To mitigate the context length limitation of LLMs and grasp spatial-temporal information in long video context, we employ a \textbf{\textit{hierarchical}} tree structure. This structure first condenses dense captions to refine event descriptions and then segments content into sequential events with finer temporal granularity. Such representations facilitate the generation of QA pairs with \textit{explicit timestamps} and varying \textit{contextual lengths}, ensuring adaptability across different temporal levels (\textit{frame level, event level, segment level}). Additionally, we design prompting strategies to generate high-quality questions that require condensed reasoning. Finally, a \textbf{\textit{self-revision}} step refines results by removing redundancy and irrelevant prior information. For more details, please refer to~\autoref{sec:dataset}.}
\label{fig:pipeline}
\end{figure*}

\section{The LongViTU Dataset}
\label{sec:dataset}

\textbf{Overview.}~~From a high level, we extract key video details, such as captions and human narration, and send them into an LLM to generate QA pairs. However, these details from long-form videos usually exceed the context limit of LLMs. Also, lengthy details make it difficult for LLM to grasp spatial-temporal information and generate favorable QA pairs (\eg, long certificate length). To address this, we employ a hierarchical tree structure: first condensing dense captions to refine event descriptions, then segmenting content into sequential events with finer temporal granularity (detailed in \autoref{subsec:videotree}). Based on the tree-based video representation, we design several prompting strategies to facilitate generating high-quality questions that require long video context and condensed reasoning (detailed in \autoref{subsec:stage2_qagen}). Finally, we conduct extensive analysis and quality assessment to ensure its quality (detailed in \autoref{subsec:char_and_stat} and \autoref{subsec:quality_assessment}).

\subsection{Dataset Construction Pipeline}
\subsubsection{Stage I: Tree-structured Video Representations}
\label{subsec:videotree}

\noindent
We design a hierarchical tree structure that encapsulates video semantics across multiple levels of granularity. This framework integrates frame-level, event-level, and segment-level, culminating in a structured tree \(\mathcal{T}_{\text{video}}\). A visual illustration can be found in  \autoref{fig:pipeline}. 

\noindent
\textbf{Frame Level.}~~Commencing at the frame level, we leverage InternLM-XComposer2~\citep{internlm-xcomposer2} to perform multi-frame dense captioning (sampled at 1 fps) across annotated events in the Ego4D~\citep{ego4d}. Descriptions are structured as $\langle d_f, t_s^f, t_e^f \rangle$, containing frame-level text description $d_f$ and explicit timestamps $t_s^f$ and $t_e^f$ derived from Ego4D's temporal annotations for each event provided by human annotators.
\begin{equation}
\mathcal{F}^k = \langle d_f^k, t_s^{f^k}, t_e^{f^k} \rangle, \quad k = 1, \dots, F
\end{equation}
where $F$ denotes the total number of sampled frames.

\noindent
\textbf{Event Level.}~~To eliminate redundant frame level text, an LLM is used to summarize both manually annotated events from Ego4D and the above frame-level automated dense captions, yielding concise event level descriptions $\langle d_e, t_s^e, t_e^e \rangle$ with explicit temporal boundaries $t_s^e$ and $t_e^e$.

\begin{equation}
\mathcal{E}^j = \langle d_e^j, t_s^{e^j}, t_e^{e^j}, \{ \mathcal{F}^k \}_{k=1}^{F} \rangle, \quad j = 1, \dots, E
\end{equation}
where $E$ denotes the total number of events.

\begin{figure*}[t!]
    \centering
    \begin{subfigure}[b]{0.59\textwidth}
        \centering
        \includegraphics[width=\textwidth]{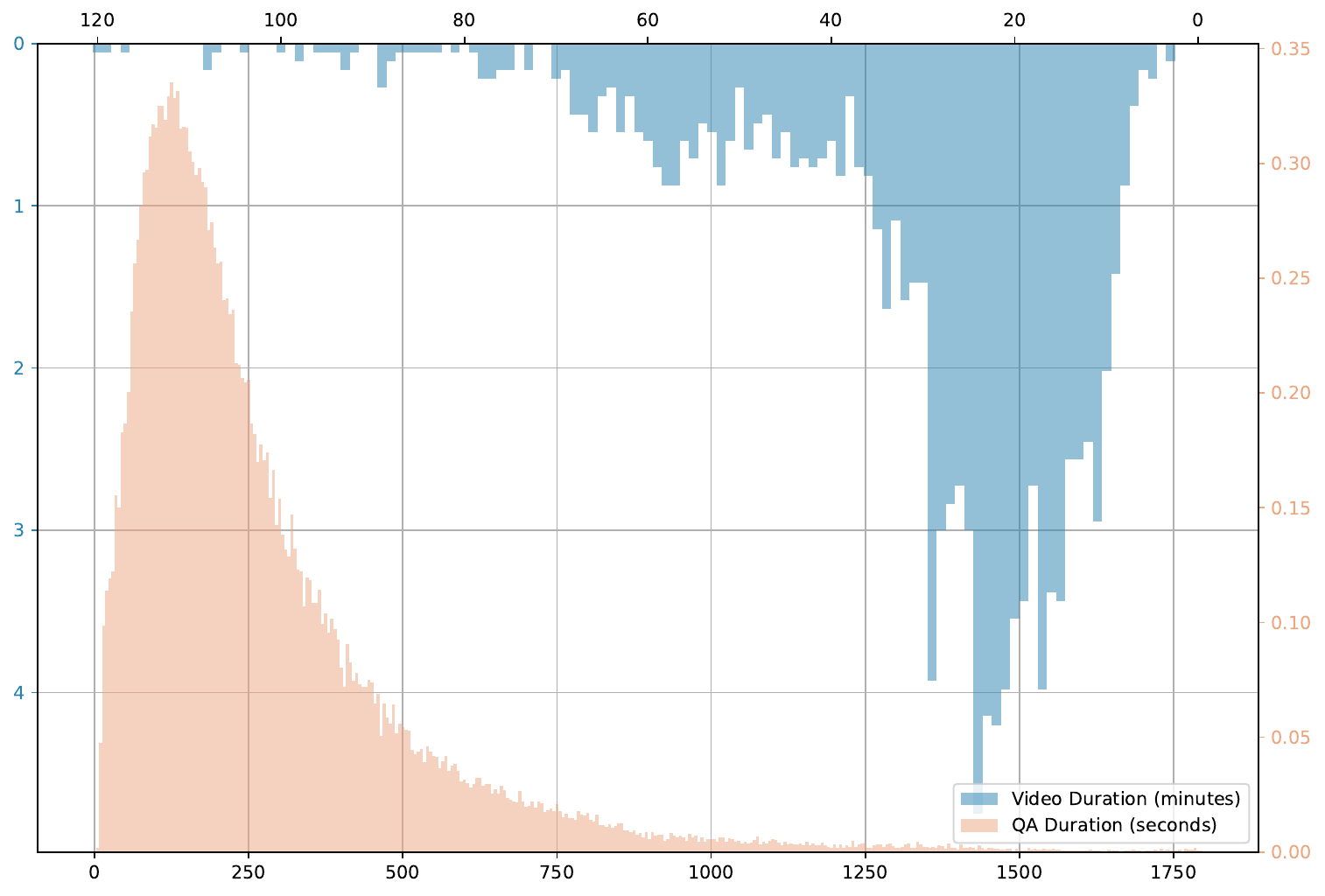}
        \caption{Video and QA duration distribution}
        \label{fig:duration}
    \end{subfigure}
    \hfill
    \begin{subfigure}[b]{0.40\textwidth}
        \centering
        \includegraphics[width=\textwidth]{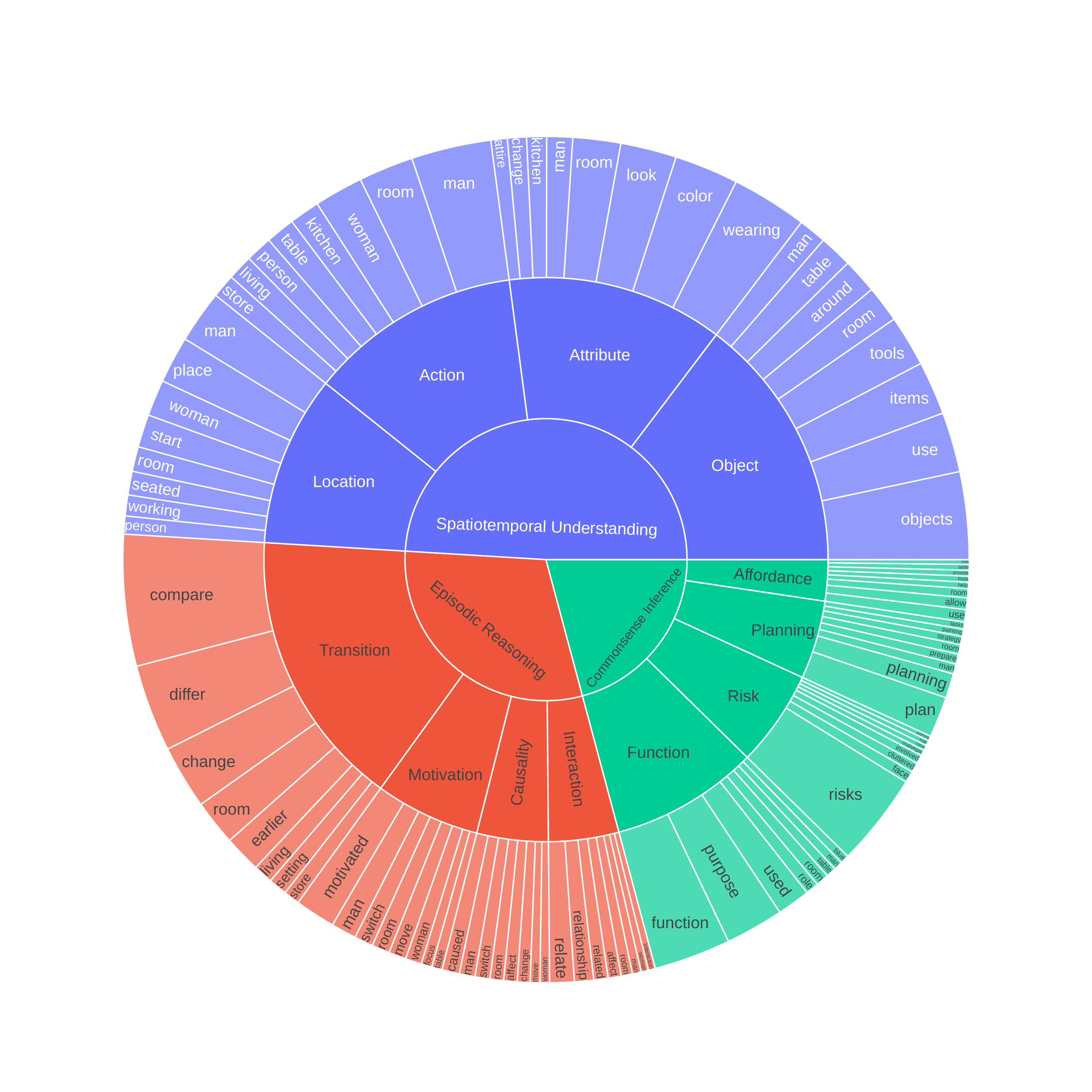}
        \caption{Sunburst of LongViTU}
        \label{fig:sunburst}
    \end{subfigure}
    \caption{\textbf{Statistics of LongViTU.}~~Subfigure (a) depicts the distributions of video and QA durations within LongViTU. The bottom horizontal axis (from left to right) represents QA duration in seconds, while the right vertical axis (from bottom to top) shows the percentage of the total dataset. QA durations predominantly vary from 6 to 600 seconds, with an average of 276.8 seconds. The top horizontal axis (from right to left) denotes video durations, and the left vertical axis (from top to bottom) indicates their percentage distribution. Video durations primarily range from 5 to 60 minutes, with an average of 29.3 minutes, following a long-tail distribution. Subfigure (b) illustrates the QA categorization in LongViTU along with their word frequency distribution. The outermost ring of the sunburst chart highlights the eight most frequent words in each category, with segment sizes reflecting their proportional frequencies. Please zoom in for a clearer view.}
    \label{fig:statistics}
\end{figure*}

\noindent
\textbf{Segment Level.}~~The LLM further organizes these events into segments in the hierarchical video tree $\mathcal{T}_{\text{video}}$, merging related events into segments summarized at the segment level, represented as $\langle d_s, t_s^s, t_e^s \rangle$.
\begin{equation}
\mathcal{S}^i = \langle d_s^i, t_s^{s^i}, t_e^{s^i}, \{ \mathcal{E}^j \}_{j=1}^{E} \rangle, \quad i = 1, \dots, S
\end{equation}
where $S$ denotes the total number of segments.

\noindent
\textbf{Video Tree.}~~Drawing on the pipeline above, we formalize the hierarchical tree structure for long-form video content as follows:
\begin{equation}
\mathcal{T}_{\text{video}} = \langle R, \{ \mathcal{S}^i \}_{i=1}^{S} \rangle
\end{equation}
where $\mathcal{T}_{\text{video}}$ represents the hierarchical tree, with $R$ as the root node and nodes down to frames.

\subsubsection{Stage II: QA Generation using Video Trees}
\label{subsec:stage2_qagen}

\noindent
With the hierarchical video representations, we fetch a sub-tree (effectively a collection of consecutive \textit{segments}) and send its corresponding video details (captions, narrations) to the LLM, then prompt it to generate the required questions based on the given video content, as detailed below.

\noindent
\textbf{Sliding Window and QA Generation.}~~A sliding window strategy over video sub-trees is employed to systematically generate QA pairs that balance spatial precision and temporal context. Specifically, each input window comprises five \textit{segments} (a hyperparameter directly influencing the average \textit{certificate length}), from which the LLM is prompted to generate QA pairs, formalized as:
\begin{equation}
\mathcal{QA}_{\text{video}} = \big\langle Q, A, \mathcal{D}_s \big\rangle,
\end{equation}
where $\mathcal{QA}_{\text{video}}$ denotes the set of generated QA pairs for the video sub-tree $\mathcal{T}_{\text{video}}$. The notation $\mathcal{D}_s$ represents the selected segments via the sliding window, where each segment $\mathcal{D}_s = \big\langle d_s^i, \mathcal{D}_e \big\rangle$ $(s=1, \dots, S)$ consists of event-level descriptions $\mathcal{D}_e = \big\langle d_e^j, \mathcal{D}_f \big\rangle$ $(e=1, \dots, E)$. Each event further includes frame-level descriptions $\mathcal{D}_f = \big\langle d_f^k \big\rangle$ $(f=1, \dots, F)$.

\noindent
\textbf{Ensuring Long Certificate Length.}~~Besides the adding aforementioned descriptions from input window of video segments, we further design the prompt to instructs the LLM to formulate questions based solely on information in the last two segments (\textbf{Ask Content}), while requiring answers exclusively found in the first three segments (\textbf{Memory Content}, see full prompt in ~\autoref{subsec:qa_generation}). This design explicitly ensures that generated questions necessitate referencing substantial preceding video context to guarante certificate lengths.

\noindent
\textbf{Facilitating Condensed Reasoning.}~~We’ve found that guiding the LLM to explicitly categorize its generated questions according to a structured reasoning taxonomy significantly improves its ability to produce questions that require condensed reasoning. Specifically, we design a hierarchical taxonomy (illustrated in \autoref{fig:sunburst}) that classifies questions by the type of reasoning involved (e.g., affordance, planning, risk, functionality, causality, etc.). During question generation, we prompt the LLM to assign each generated question to the appropriate category within this taxonomy (detailed in \autoref{subsec:definitions}). This structured categorization encourages the model to focus on generating questions that inherently demand more condensed and targeted reasoning.




\noindent
\textbf{Self-Revision.}~~After generating the initial QA pairs, the LLM undergoes rigorous self-revision to ensure consistency. This includes correcting inaccuracies, removing extraneous information, and refining responses to reduce redundancy and hallucinations, thereby improving dataset reliability. Additionally, the LLM conducts a pure-text evaluation (\textit{without visual input}) to detect and discard QA pairs with excessive textual bias. The full prompt can be found in~\autoref{subsec:selfrevision}.


\subsection{Characteristics and Statistics}
\label{subsec:char_and_stat}

\noindent
\textbf{Duration Distribution.}~~The LongViTU dataset contains 1,833 videos, divided into 1,533 for training, 200 for validation, and 100 for testing, totaling approximately 900 hours. The average video duration is 29.3 minutes, ranging from 3.5 to 120.7 minutes with a standard deviation of 17.5 minutes, following a long-tail distribution (\autoref{fig:duration}). Each video is accompanied by an average of 66 QA pairs; QA durations range from 6 to 1800 seconds, averaging 276.8 seconds, also exhibiting a long-tail pattern. Events and segments have average durations of 8.5 and 82 seconds, respectively. The dataset includes 121k QA pairs: 101k for training, 14k for validation, and 600 human-reviewed samples for testing.

\noindent
\textbf{Category Distribution.}~~The sunburst diagram in~\autoref{fig:sunburst} illustrates the distribution of QA pairs across the aforementioned hierarchical taxonomy designed by us; detailed categorization and examples are provided in~\autoref{sec:examples}.


\subsection{Quality Assessment}
\label{subsec:quality_assessment}

\begin{figure}[htbp]
    \centering
    \includegraphics[width=0.47\textwidth]{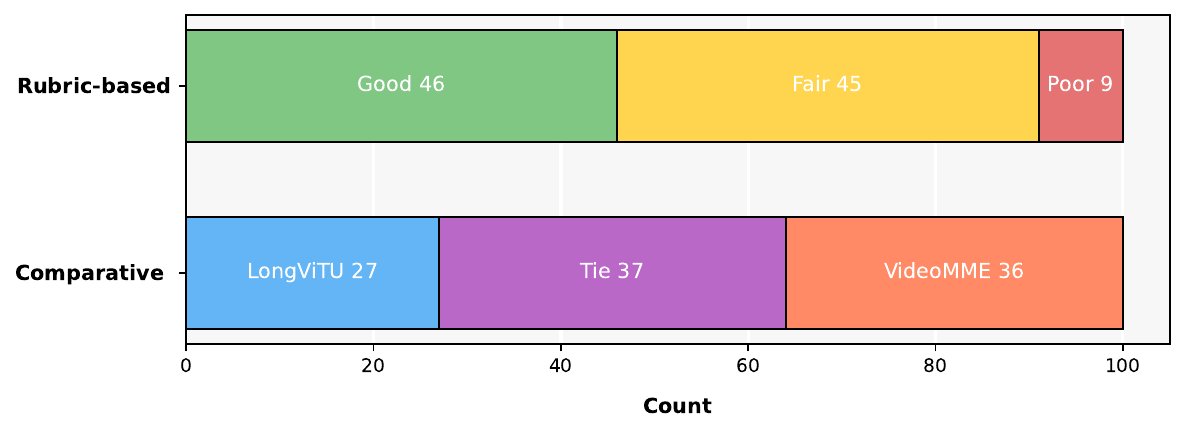}
    \caption{\textbf{Quality Assessment.}~~Rubric-based assessment of LongViTU and comparative assessment against VideoMME~\cite{videomme}.}
    \label{fig:human_study}
\end{figure}

\noindent
We recruit a pool of raters to assess the quality of LongViTU data from two aspects: 1) evaluate the quality of LongViTU questions using a predefined rubric; 2) compare the quality of automatically generated LongViTU against VideoMME~\cite{videomme}, a video QA benchmark that is annotated exclusively by humans. The results are shown in~\autoref{fig:human_study}. 

\noindent
\textbf{Rubric-based assessment.}~~For the first aspect, the raters are guided to score the questions using the following rubric:
\begin{itemize}
    \item \textbf{Good}: The video segment contains rich and diverse events, and the generated QA pair exhibits strong spatio-temporal coherence with at least one event.
    \item \textbf{Fair}: The video segment is relatively simple, and while the QA pair is well-aligned with an event, it primarily captures either spatial or temporal aspects rather than both.
    \item \textbf{Poor}: The QA pair does not align well with the video content (e.g., it could be answered without reference to the video or is overly generic). However, it does not contain factual errors or hallucinations.
\end{itemize}
Due to resource constraints, only 100 randomly selected questions from LongViTU are evaluated. As suggested by the rubric, \texttt{Good} denotes excellent questions, while \texttt{Fair} questions remain highly valuable for long-video QA. Even \texttt{Poor} does not indicate incorrect QA but rather reflects weaker video alignment, which can still contribute to training more robust models. Overall, approximately 46\% of the QA pairs exhibit strong spatio-temporal coherence, validating the quality of LongViTU. 

\noindent
\textbf{Comparative Assessment.}~~We randomly sample 100 questions each from our autonomously generated LongViTU dataset and the manually annotated VideoMME dataset. Evaluators are asked to determine whether the quality of each question from \texttt{LongViTU} is significantly better, significantly worse, or comparable (\texttt{Tie}) to its counterpart in \texttt{VideoMME}. Results indicate that approximately 64\% of questions from LongViTU achieve comparable quality to VideoMME, demonstrating that our fully autonomously generated dataset closely approaches human-annotated quality.

\section{Experiments}
\label{experiments}

In our experiments, we aim to answer the following:
\begin{enumerate}
    \item Do LongViTU questions indeed impose substantial challenges (long-video context and condensed reasoning) to today's video understanding models?
    \item To which degree does training on LongViTU enhance the capability of mainstream Vision-Language Models (VLMs) to understand long videos?
\end{enumerate}
To answer the first question, we primarily perform \textbf{zero-shot evaluation} of mainstream video understanding models on the LongViTU test set. To answer the second question, we first perform \textbf{supervised fine-tuning (SFT)} of a collection of these models on LongViTU, then evaluate them on a collection of long-video understanding benchmarks (including LongViTU test set). To comprehensively evaluate this capability, we consider both In-Distribution (ID) and several canonical Out-of-Distribution (OOD) benchmarks.\footnote{ID benchmarks share the same video source (Ego4D) as LongViTU, whereas OOD benchmarks utilize alternative sources.} Detailed experimental settings are described in \autoref{subsec:exp_setting}, and results across various benchmarks are presented in \autoref{subsec:exp_longvitu} and \autoref{subsec:exp_others}.

\begin{table*}
    \tiny
    \centering
    \caption{\textbf{Quantitative results on LongViTU.}~~All results are derived from evaluations conducted by GPT-4~\citep{gpt4}, the criteria and prompt are detailed in~\autoref{subsec:metrics}.~$^{*}$ denotes results obtained in a \colorbox{gray!20}{zero-shot} manner, while $^{**}$ indicates \colorbox{green!20}{fine-tuned} results following training on the LongViTU training set, \colorbox{blue!20}{$\triangle~\text{compared}$} highlighting the percentage difference in performance between their. "Overall Avg." represents the average scores across three primary categories. The top-performing open-source model, LongVU~\citep{longvu}, achieved a score of 55.9, surpassing the 52.3 score of the best proprietary model, Gemini-1.5-pro~\citep{geminipro}, these results remain far from human-level capabilities.}
    \resizebox{0.95\textwidth}{!}{
    \begin{tabular}{llcccccc}
    \toprule
    
    \multirow{2}{*}{\textbf{Setting}} & \multirow{2}{*}{\textbf{Method}} & \multirow{2}{*}{\textbf{Overall Avg.}} & \multicolumn{5}{c}{\textbf{Spatiotemporal Understanding}} \\ \cmidrule{4-8} & & & Object & Attribute & Location & Action & Avg. \\
    \midrule
    \multirow{1}{*}{\textbf{Blind}} & GPT-4 turbo & 38.2 & 26.1 & 33.2 & 32.0 & 29.4 & 30.2 \\
    \midrule
    \multirow{3}{*}{\textbf{Frame-Based}}
    & \cellcolor{gray!20}Video-LLaVA$^{*}$ & \cellcolor{gray!20}45.9 & \cellcolor{gray!20}37.8 & \cellcolor{gray!20}46.3 & \cellcolor{gray!20}49.1 & \cellcolor{gray!20}38.1 & \cellcolor{gray!20}42.7 \\
    & \cellcolor{green!20}Video-LLaVA$^{**}$ & \cellcolor{green!20}50.7 & \cellcolor{green!20}39.3 & \cellcolor{green!20}49.2 & \cellcolor{green!20}49.6 & \cellcolor{green!20}41.8 & \cellcolor{green!20}44.9 \\
    & \cellcolor{blue!20}$\triangle~\text{compared}$ & \cellcolor{blue!20}+10.5\% & \cellcolor{blue!20}+4.0\% & \cellcolor{blue!20}+6.3\% & \cellcolor{blue!20}+1.0\% & \cellcolor{blue!20}+9.7\% & \cellcolor{blue!20}+5.2\% \\
    \midrule
    \multirow{10}{*}{\textbf{Sampling-Based}}
    & VideoAgent$^{*}$ & 44.0 & 35.7 & 43.1 & 45.9 & 36.4 & 40.2 \\
    & \cellcolor{gray!20}LLaMA-VID$^{*}$ & \cellcolor{gray!20}38.2 & \cellcolor{gray!20}29.4 & \cellcolor{gray!20}35.6 & \cellcolor{gray!20}40.1 & \cellcolor{gray!20}31.5 & \cellcolor{gray!20}34.3 \\
    & \cellcolor{green!20}LLaMA-VID$^{**}$ & \cellcolor{green!20}44.5 & \cellcolor{green!20}33.5 & \cellcolor{green!20}37.4 & \cellcolor{green!20}45.7 & \cellcolor{green!20}37.6 & \cellcolor{green!20}39.1 \\
    & \cellcolor{blue!20}$\triangle~\text{compared}$ & \cellcolor{blue!20}+16.5\% & \cellcolor{blue!20}+13.9\% & \cellcolor{blue!20}+5.1\% & \cellcolor{blue!20}+14.0\% & \cellcolor{blue!20}+19.4\% & \cellcolor{blue!20}+14.0\% \\
    & LongVA-DPO$^{*}$ & 47.5 & 35.9 & 52.4 & 44.3 & 37.2 & 41.8 \\
    & \cellcolor{gray!20}LongVU$^{*}$ & \cellcolor{gray!20}49.9 & \cellcolor{gray!20}39.3 & \cellcolor{gray!20}47.6 & \cellcolor{gray!20}52.3 & \cellcolor{gray!20}44.3 & \cellcolor{gray!20}46.3 \\
    & \cellcolor{green!20}LongVU$^{**}$ & \cellcolor{green!20}55.9 & \cellcolor{green!20}40.2 & \cellcolor{green!20}55.2 & \cellcolor{green!20}54.3 & \cellcolor{green!20}45.3 & \cellcolor{green!20}48.8 \\
    & \cellcolor{blue!20}$\triangle~\text{compared}$ & \cellcolor{blue!20}+12.0\% & \cellcolor{blue!20}+2.3\% & \cellcolor{blue!20}+16.0\% & \cellcolor{blue!20}+3.8\% & \cellcolor{blue!20}+2.3\% & \cellcolor{blue!20}+5.4\% \\
    & Gemini-1.5-Pro$^{*}$ & 52.3 & 54.3 & 58.6 & 56.3 & 48.1 & 54.7 \\
    & \cellcolor{gray!50}Human & \cellcolor{gray!50}81.0 & \cellcolor{gray!50}88.3 & \cellcolor{gray!50}82.0 & \cellcolor{gray!50}86.9 & \cellcolor{gray!50}85.8 & \cellcolor{gray!50}84.1 \\
    \midrule
    
    \multirow{2}{*}{\textbf{Setting}} & \multirow{2}{*}{\textbf{Method}} & \multirow{2}{*}{\textbf{Overall Avg.}} & \multicolumn{5}{c}{\textbf{Episodic Reasoning}} \\ \cmidrule{4-8} & & & Transition & Interaction & Causality & Motivation & Avg. \\
    \midrule
    \multirow{1}{*}{\textbf{Blind}} & GPT-4 turbo & 38.2 & 45.1 & 47.4 & 47.7 & 56.1 & 49.5 \\
    \midrule
    \multirow{3}{*}{\textbf{Frame-Based}}
    & \cellcolor{gray!20}Video-LLaVA$^{*}$ & \cellcolor{gray!20}45.9 & \cellcolor{gray!20}45.6 & \cellcolor{gray!20}50.5 & \cellcolor{gray!20}48.8 & \cellcolor{gray!20}53.2 & \cellcolor{gray!20}49.4 \\
    & \cellcolor{green!20}Video-LLaVA$^{**}$ & \cellcolor{green!20}50.7 & \cellcolor{green!20}50.5 & \cellcolor{green!20}56.4 & \cellcolor{green!20}59.7 & \cellcolor{green!20}64.9 & \cellcolor{green!20}58.0 \\
    & \cellcolor{blue!20}$\triangle~\text{compared}$ & \cellcolor{blue!20}+10.5\% & \cellcolor{blue!20}+10.7\% & \cellcolor{blue!20}+11.7\% & \cellcolor{blue!20}+22.3\% & \cellcolor{blue!20}+22.0\% & \cellcolor{blue!20}+17.4\% \\
    \midrule
    \multirow{10}{*}{\textbf{Sampling-Based}}
    & VideoAgent$^{*}$ & 44.0 & 43.1 & 45.5 & 49.9 & 52.8 & 48.1 \\
    & \cellcolor{gray!20}LLaMA-VID$^{*}$ & \cellcolor{gray!20}38.2 & \cellcolor{gray!20}40.4 & \cellcolor{gray!20}46.7 & \cellcolor{gray!20}40.5 & \cellcolor{gray!20}46.6 & \cellcolor{gray!20}43.2 \\
    & \cellcolor{green!20}LLaMA-VID$^{**}$ & \cellcolor{green!20}44.5 & \cellcolor{green!20}46.7 & \cellcolor{green!20}48.4 & \cellcolor{green!20}54.2 & \cellcolor{green!20}57.7 & \cellcolor{green!20}52.1 \\
    & \cellcolor{blue!20}$\triangle~\text{compared}$ & \cellcolor{blue!20}+16.5\% & \cellcolor{blue!20}+15.6\% & \cellcolor{blue!20}+3.6\% & \cellcolor{blue!20}+33.8\% & \cellcolor{blue!20}+23.8\% & \cellcolor{blue!20}+20.6\% \\
    & LongVA-DPO$^{*}$ & 47.5 & 51.8 & 54.6 & 52.2 & 59.5 & 54.8 \\
    & \cellcolor{gray!20}LongVU$^{*}$ & \cellcolor{gray!20}49.9 & \cellcolor{gray!20}56.7 & \cellcolor{gray!20}58.8 & \cellcolor{gray!20}48.9 & \cellcolor{gray!20}53.4 & \cellcolor{gray!20}54.2 \\
    & \cellcolor{green!20}LongVU$^{**}$ & \cellcolor{green!20}55.9 & \cellcolor{green!20}62.4 & \cellcolor{green!20}63.5 & \cellcolor{green!20}63.2 & \cellcolor{green!20}70.0 & \cellcolor{green!20}65.2 \\
    & \cellcolor{blue!20}$\triangle~\text{compared}$ & \cellcolor{blue!20}+12.0\% & \cellcolor{blue!20}+10.1\% & \cellcolor{blue!20}+8.0\% & \cellcolor{blue!20}+29.2\% & \cellcolor{blue!20}+31.1\% & \cellcolor{blue!20}+20.3\% \\
    & Gemini-1.5-Pro$^{*}$ & 52.3 & 47.8 & 45.5 & 47.8 & 47.5 & 47.3 \\
    & \cellcolor{gray!50}Human & \cellcolor{gray!50}81.0 & \cellcolor{gray!50}62.4 & \cellcolor{gray!50}80.5 & \cellcolor{gray!50}78.2 & \cellcolor{gray!50}76.2 & \cellcolor{gray!50}74.3 \\
    \midrule
    
    \multirow{2}{*}{\textbf{Setting}} & \multirow{2}{*}{\textbf{Method}} & \multirow{2}{*}{\textbf{Overall Avg.}} & \multicolumn{5}{c}{\textbf{Commonsense Inference}} \\ \cmidrule{4-8} & & & Planning & Risk & Function & Affordance & Avg. \\
    \midrule
    \multirow{1}{*}{\textbf{Blind}} & GPT-4 turbo & 38.2 & 36.5 & 51.1 & 55.9 & 50.9 & 48.7 \\
    \midrule
    \multirow{3}{*}{\textbf{Frame-Based}}
    & \cellcolor{gray!20}Video-LLaVA$^{*}$ & \cellcolor{gray!20}45.9 & \cellcolor{gray!20}41.6 & \cellcolor{gray!20}56.8 & \cellcolor{gray!20}55.3 & \cellcolor{gray!20}54.6 & \cellcolor{gray!20}51.7 \\
    & \cellcolor{green!20}Video-LLaVA$^{**}$ & \cellcolor{green!20}50.7 & \cellcolor{green!20}50.2 & \cellcolor{green!20}62.6 & \cellcolor{green!20}64.0 & \cellcolor{green!20}64.6 & \cellcolor{green!20}59.8 \\
    & \cellcolor{blue!20}$\triangle~\text{compared}$ & \cellcolor{blue!20}+10.5\% & \cellcolor{blue!20}+20.7\% & \cellcolor{blue!20}+10.2\% & \cellcolor{blue!20}+15.7\% & \cellcolor{blue!20}+18.3\% & \cellcolor{blue!20}+15.7\% \\
    \midrule
    \multirow{10}{*}{\textbf{Sampling-Based}}
    & VideoAgent$^{*}$ & 44.0 & 40.0 & 53.7 & 55.5 & 53.1 & 50.7 \\
    & \cellcolor{gray!20}LLaMA-VID$^{*}$ & \cellcolor{gray!20}38.2 & \cellcolor{gray!20}34.9 & \cellcolor{gray!20}51.3 & \cellcolor{gray!20}46.5 & \cellcolor{gray!20}47.2 & \cellcolor{gray!20}44.1 \\
    & \cellcolor{green!20}LLaMA-VID$^{**}$ & \cellcolor{green!20}44.5 & \cellcolor{green!20}43.9 & \cellcolor{green!20}54.5 & \cellcolor{green!20}55.7 & \cellcolor{green!20}53.8 & \cellcolor{green!20}51.7 \\
    & \cellcolor{blue!20}$\triangle~\text{compared}$ & \cellcolor{blue!20}+16.5\% & \cellcolor{blue!20}+25.8\% & \cellcolor{blue!20}+6.2\% & \cellcolor{blue!20}+19.8\% & \cellcolor{blue!20}+14.0\% & \cellcolor{blue!20}+17.2\% \\
    & LongVA-DPO$^{*}$ & 47.5 & 46.3 & 57.0 & 61.9 & 59.2 & 56.1 \\
    & \cellcolor{gray!20}LongVU$^{*}$ & \cellcolor{gray!20}49.9 & \cellcolor{gray!20}46.2 & \cellcolor{gray!20}52.3 & \cellcolor{gray!20}60.4 & \cellcolor{gray!20}62.8 & \cellcolor{gray!20}56.0 \\
    & \cellcolor{green!20}LongVU$^{**}$ & \cellcolor{green!20}55.9 & \cellcolor{green!20}59.9 & \cellcolor{green!20}67.9 & \cellcolor{green!20}68.7 & \cellcolor{green!20}70.4 & \cellcolor{green!20}66.4 \\
    & \cellcolor{blue!20}$\triangle~\text{compared}$ & \cellcolor{blue!20}+12.0\% & \cellcolor{blue!20}+29.7\% & \cellcolor{blue!20}+29.8\% & \cellcolor{blue!20}+13.7\% & \cellcolor{blue!20}+12.1\% & \cellcolor{blue!20}+18.6\% \\
    & Gemini-1.5-Pro$^{*}$ & 52.3 & 43.6 & 57.5 & 46.1 & 43.6 & 50.3 \\
    & \cellcolor{gray!50}Human & \cellcolor{gray!50}81.0 & \cellcolor{gray!50}70.2 & \cellcolor{gray!50}78.6 & \cellcolor{gray!50}78.7 & \cellcolor{gray!50}69.0 & \cellcolor{gray!50}75.5 \\
    
    \bottomrule
    \end{tabular}}
    \label{tab:longvitu}
\end{table*}

\begin{table*}
    \tiny
    \centering
    \caption{\textbf{Quantitative results on additional benchmarks.}~~The~$^{*}$ denotes results obtained in a \colorbox{gray!20}{zero-shot} manner, while $^{**}$ indicates \colorbox{green!20}{fine-tuned} results following training on the LongViTU training set, \colorbox{blue!20}{$\triangle~\text{compared}$} highlighting the percentage difference in performance between them. The zero-shot results of SFT with LongViTU on Egoschema~\citep{egoschema}, VideoMME~\citep{videomme}, MLVU~\citep{mlvu}, LVBench~\citep{lvbench}, and MVBench~\citep{mvbench} are reproduced from the official checkpoint on Hugging Face (\href{https://huggingface.co/Vision-CAIR/LongVU_Qwen2_7B}{LongVU}, \href{https://huggingface.co/lmms-lab/LLaVA-Video-7B-Qwen2}{LLaVA-Video}), rather than those reported in the arXiv (\href{https://arxiv.org/pdf/2410.17434}{LongVU}, \href{https://arxiv.org/pdf/2410.02713}{LLaVA-Video}). MVBench~\citep{mvbench} only shows the subc-category that have shown improvement.}
    \resizebox{1\textwidth}{!}{
    \begin{tabular}{l|c|c|c|c|c|c|c|c|c}
    \toprule
    \multirow{2}{*}{\textbf{Method}} & \multirow{2}{*}{\textbf{EgoSchema}} & \multicolumn{4}{c|}{\textbf{VideoMME}} & \multirow{2}{*}{\textbf{WorldQA}} & \multicolumn{3}{c}{\textbf{OpenEQA}} \\
    \cmidrule{3-6}
    \cmidrule{8-10}
    & & Avg. & Short & Medium & Long & & Avg. & ScanNet & HM3D \\
    \midrule
    
    VideoLLM Online$^{*}$ & 47.4 & 13.7 & 24.3 & 16.7 & 0.0 & 30.0 & 23.3 & 24.8 & 20.4 \\
    \midrule
    
    \cellcolor{gray!20}LLaMA-VID$^{*s_3}$ & \cellcolor{gray!20}23.6 & \cellcolor{gray!20}14.6 & \cellcolor{gray!20}19.5 & \cellcolor{gray!20}12.6 & \cellcolor{gray!20}11.5 & \cellcolor{gray!20}30.9 & \cellcolor{gray!20}31.1 & \cellcolor{gray!20}31.0 & \cellcolor{gray!20}31.3 \\
    \cellcolor{gray!20}LLaMA-VID$^{*s_2}$ & \cellcolor{gray!20}30.4 & \cellcolor{gray!20}16.7 & \cellcolor{gray!20}22.6 & \cellcolor{gray!20}15.3 & \cellcolor{gray!20}12.2 & \cellcolor{gray!20}32.0 & \cellcolor{gray!20}31.9 & \cellcolor{gray!20}31.8 & \cellcolor{gray!20}32.1 \\
    \cellcolor{green!20}LLaMA-VID$^{**}$ & \cellcolor{green!20}34.0 & \cellcolor{green!20}17.2 & \cellcolor{green!20}23.8 & \cellcolor{green!20}15.4 & \cellcolor{green!20}12.2 & \cellcolor{green!20}32.2 & \cellcolor{green!20}33.6 & \cellcolor{green!20}33.5 & \cellcolor{green!20}33.8\\
    \cellcolor{blue!20}$\triangle~\text{compared}$ & \cellcolor{blue!20}+11.8\% & \cellcolor{blue!20}+3.0\% & \cellcolor{blue!20}+5.3\% & \cellcolor{blue!20}+0.7\% & \cellcolor{blue!20}+0.0\% & \cellcolor{blue!20}+0.6\% & \cellcolor{blue!20}+5.3\% & \cellcolor{blue!20}+5.3\% & \cellcolor{blue!20}+5.3\% \\
    \midrule
    
    \cellcolor{gray!20}Video-LLaVA$^{*}$ & \cellcolor{gray!20}36.8 & \cellcolor{gray!20}32.3 & \cellcolor{gray!20}33.7 & \cellcolor{gray!20}31.6 & \cellcolor{gray!20}31.5 & \cellcolor{gray!20}30.2 & \cellcolor{gray!20}35.1 & \cellcolor{gray!20}37.3 & \cellcolor{gray!20}30.9 \\
    \cellcolor{green!20}Video-LLaVA$^{**}$ & \cellcolor{green!20}48.1 & \cellcolor{green!20}32.5 & \cellcolor{green!20}30.5 & \cellcolor{green!20}33.7 & \cellcolor{green!20}33.1 & \cellcolor{green!20}34.1 & \cellcolor{green!20}32.6 & \cellcolor{green!20}32.6 & \cellcolor{green!20}32.5 \\
    \cellcolor{blue!20}$\triangle~\text{compared}$ & \cellcolor{blue!20}+30.7\%& \cellcolor{blue!20}+0.6\% & \cellcolor{blue!20}\textcolor{red}{-9.5\%} & \cellcolor{blue!20}+6.6\% & \cellcolor{blue!20}+5.1\% & \cellcolor{blue!20}+12.9\% & \cellcolor{blue!20}\textcolor{red}{-7.1\%} & \cellcolor{blue!20}\textcolor{red}{-12.6\%} & \cellcolor{blue!20}+5.2\% \\
    \midrule

    LongVA-DPO$^{*}$ & 56.9 & 54.3 & 61.6 & 50.4 & 47.6 & 30.3 & 36.6 & 41.5 & 26.9 \\
    \cellcolor{gray!20}LongVU$^{*}$ & \cellcolor{gray!20}67.6 & \cellcolor{gray!20}55.6 & \cellcolor{gray!20}66.0 & \cellcolor{gray!20}54.1 & \cellcolor{gray!20}46.6 & \cellcolor{gray!20}35.7 & \cellcolor{gray!20}48.3 & \cellcolor{gray!20}51.1 & \cellcolor{gray!20}42.8 \\
    \cellcolor{green!20}LongVU$^{**}$ & \cellcolor{green!20}70.8 & \cellcolor{green!20}55.9 & \cellcolor{green!20}66.2 & \cellcolor{green!20}54.3 & \cellcolor{green!20}47.0 & \cellcolor{green!20}36.5 & \cellcolor{green!20}48.9 & \cellcolor{green!20}51.4 & \cellcolor{green!20}44.2 \\
    \cellcolor{blue!20}$\triangle~\text{compared}$ & \cellcolor{blue!20}+4.7\%& \cellcolor{blue!20}+0.5\% & \cellcolor{blue!20}+0.3\% & \cellcolor{blue!20}+0.4\% & \cellcolor{blue!20}+0.9\% & \cellcolor{blue!20}+2.2\% & \cellcolor{blue!20}+1.2\% & \cellcolor{blue!20}+0.6\% & \cellcolor{blue!20}+3.3\% \\
    \midrule

    \cellcolor{gray!20}LLaVA-Video$^{*}$ & \cellcolor{gray!20}57.3 & \cellcolor{gray!20}61.9 & \cellcolor{gray!20}74.0 & \cellcolor{gray!20}60.2 & \cellcolor{gray!20}51.6 & \cellcolor{gray!20}- & \cellcolor{gray!20}- & \cellcolor{gray!20}- & \cellcolor{gray!20}- \\
    \cellcolor{green!20}LLaVA-Video$^{**}$ & \cellcolor{green!20}62.8 & \cellcolor{green!20}62.2 & \cellcolor{green!20}73.9 & \cellcolor{green!20}60.4 & \cellcolor{green!20}52.1 & \cellcolor{green!20}- & \cellcolor{green!20}- & \cellcolor{green!20}- & \cellcolor{green!20}- \\
    \cellcolor{blue!20}$\triangle~\text{compared}$ & \cellcolor{blue!20}+9.6\%& \cellcolor{blue!20}{+0.4\%} & \cellcolor{blue!20}\textcolor{red}{-0.1\%} & \cellcolor{blue!20}{+0.3\%} & \cellcolor{blue!20}+1.0\% & \cellcolor{blue!20}- & \cellcolor{blue!20}- & \cellcolor{blue!20}- & \cellcolor{blue!20}- \\
    
    \bottomrule
    \end{tabular}}

    \vspace{3mm} 

    \resizebox{1\textwidth}{!}{
    \begin{tabular}{l|c|c|c|c|c|c|c|c}
    \toprule
    \multirow{2}{*}{\textbf{Method}} & \multicolumn{8}{c}{\textbf{MLVU}} \\
    \cmidrule{2-9}
    & \textbf{Count} & \textbf{Ego} & \textbf{Needle} & \textbf{Order} & \textbf{PlotQA} & \textbf{Anomaly Reco.} & \textbf{Topic Reasoning} & \textbf{Avg.} \\
    \midrule
    
    \cellcolor{gray!20}LongVU$^{*}$ & \cellcolor{gray!20}27.9 & \cellcolor{gray!20}55.7 & \cellcolor{gray!20}73.5 & \cellcolor{gray!20}53.3 & \cellcolor{gray!20}68.5 & \cellcolor{gray!20}74.5 & \cellcolor{gray!20}86.7 & \cellcolor{gray!20}62.9 \\
    \cellcolor{green!20}LongVU$^{**}$ & \cellcolor{green!20}29.8 & \cellcolor{green!20}62.5 & \cellcolor{green!20}75.5 & \cellcolor{green!20}54.8 & \cellcolor{green!20}68.8 & \cellcolor{green!20}73.5 & \cellcolor{green!20}86.7 & \cellcolor{green!20}65.2 \\
    \cellcolor{blue!20}$\triangle~\text{compared}$ & \cellcolor{blue!20}+6.8\% & \cellcolor{blue!20}+12.2\% & \cellcolor{blue!20}+2.7\% & \cellcolor{blue!20}+2.8\% & \cellcolor{blue!20}+0.4\% & \cellcolor{blue!20}\textcolor{red}{-1.3\%} & \cellcolor{blue!20}+0.0\% & \cellcolor{blue!20}+3.7\% \\
    \midrule
    \cellcolor{gray!20}LLaVA-Video$^{*}$ & \cellcolor{gray!20}38.5 & \cellcolor{gray!20}58.2 & \cellcolor{gray!20}75.5 & \cellcolor{gray!20}65.3 & \cellcolor{gray!20}77.0 & \cellcolor{gray!20}74.0 & \cellcolor{gray!20}85.2 & \cellcolor{gray!20}67.7 \\
    \cellcolor{green!20}LLaVA-Video$^{**}$ & \cellcolor{green!20}39.4 & \cellcolor{green!20}63.4 & \cellcolor{green!20}76.1 & \cellcolor{green!20}66.8 & \cellcolor{green!20}77.0 & \cellcolor{green!20}74.0 & \cellcolor{green!20}86.7 & \cellcolor{green!20}69.9 \\
    \cellcolor{blue!20}$\triangle~\text{compared}$ & \cellcolor{blue!20}+2.3\% & \cellcolor{blue!20}+8.9\% & \cellcolor{blue!20}+0.8\% & \cellcolor{blue!20}+2.3\% & \cellcolor{blue!20}+0.0\% & \cellcolor{blue!20}+0.0\% & \cellcolor{blue!20}+1.8\% & \cellcolor{blue!20}+3.2\% \\
    
    \bottomrule
    \end{tabular}}

    \vspace{3mm} 

    \resizebox{1\textwidth}{!}{
    \begin{tabular}{l|c|c|c|c|c|c|c}
    \toprule
    \multirow{2}{*}{\textbf{Method}} & \multicolumn{7}{c}{\textbf{LVBench}} \\
    \cmidrule{2-8}
    & \textbf{Key Info.} & \textbf{Entity Recog.} & \textbf{Event Underst.} & \textbf{Reason.} & \textbf{Temp. Ground.} & \textbf{Summar.} & \textbf{Avg.} \\
    \midrule
    
    \cellcolor{gray!20}LongVU$^{*}$ & \cellcolor{gray!20}38.1 & \cellcolor{gray!20}38.4 & \cellcolor{gray!20}34.2 & \cellcolor{gray!20}43.5 & \cellcolor{gray!20}27.7 & \cellcolor{gray!20}28.6 & \cellcolor{gray!20}35.1 \\
    \cellcolor{green!20}LongVU$^{**}$ & \cellcolor{green!20}38.1 & \cellcolor{green!20}39.8 & \cellcolor{green!20}35.5 & \cellcolor{green!20}44.2 & \cellcolor{green!20}28.8 & \cellcolor{green!20}26.8 & \cellcolor{green!20}35.3 \\
    \cellcolor{blue!20}$\triangle~\text{compared}$ & \cellcolor{blue!20}+0.0\% & \cellcolor{blue!20}+3.6\% & \cellcolor{blue!20}+3.8\% & \cellcolor{blue!20}+1.6\% & \cellcolor{blue!20}+4.0\% & \cellcolor{blue!20}\textcolor{red}{-6.3\%} & \cellcolor{blue!20}+0.6\% \\
    \midrule
    \cellcolor{gray!20}LLaVA-Video$^{*}$ & \cellcolor{gray!20}45.2 & \cellcolor{gray!20}42.3 & \cellcolor{gray!20}41.4 & \cellcolor{gray!20}48.3 & \cellcolor{gray!20}33.8 & \cellcolor{gray!20}26.8 & \cellcolor{gray!20}39.6 \\
    \cellcolor{green!20}LLaVA-Video$^{**}$ & \cellcolor{green!20}45.2 & \cellcolor{green!20}42.7 & \cellcolor{green!20}41.2 & \cellcolor{green!20}49.0 & \cellcolor{green!20}35.4 & \cellcolor{green!20}26.8 & \cellcolor{green!20}40.0 \\
    \cellcolor{blue!20}$\triangle~\text{compared}$ & \cellcolor{blue!20}+0.0\% & \cellcolor{blue!20}+0.9\% & \cellcolor{blue!20}\textcolor{red}{-0.5}\% & \cellcolor{blue!20}+1.4\% & \cellcolor{blue!20}+4.7\% & \cellcolor{blue!20}+0.0\% & \cellcolor{blue!20}+1.0\% \\
    
    \bottomrule
    \end{tabular}}

    \vspace{3mm} 

    \resizebox{1\textwidth}{!}{
    \begin{tabular}{l|c|c|c|c|c|c}
    \toprule
    \multirow{2}{*}{\textbf{Method}} & \multicolumn{6}{c}{\textbf{MVBench}} \\
    \cmidrule{2-7}
    & \textbf{Action Pred.} & \textbf{Episodic Reason.} & \textbf{Moving Count} & \textbf{Moving Dir.} & \textbf{Object Shuffle} & \textbf{Unexp. Action} \\
    \midrule
    
    \cellcolor{gray!20}LongVU$^{*}$ & \cellcolor{gray!20}66.5 & \cellcolor{gray!20}52.5 & \cellcolor{gray!20}79.0 & \cellcolor{gray!20}90.0 & \cellcolor{gray!20}38.5 & \cellcolor{gray!20}72.5 \\
    \cellcolor{green!20}LongVU$^{**}$ & \cellcolor{green!20}67.0 & \cellcolor{green!20}55.5 & \cellcolor{green!20}80.0 & \cellcolor{green!20}90.5 & \cellcolor{green!20}39.5 & \cellcolor{green!20}77.5 \\
    \cellcolor{blue!20}$\triangle~\text{compared}$ & \cellcolor{blue!20}+0.8\% & \cellcolor{blue!20}+5.7\% & \cellcolor{blue!20}+1.3\% & \cellcolor{blue!20}+0.6\% & \cellcolor{blue!20}+2.6\% & \cellcolor{blue!20}+6.9\% \\
    
    \bottomrule
    \end{tabular}}
    \label{tab:ood}
\end{table*}%

\subsection{Setup}
\label{subsec:exp_setting}

\noindent
\textbf{Settings and Models.}~~We consider a variety of video understanding models, including frame-sampling-based, long-context (therefore, no need to do frame sampling), and multimodal agents.
For frame-sampling-based models such as Video-LLaVA~\citep{videollava}, we uniformly sample 8 frames per video as input. For long-context LLaMA-VID~\citep{llamavid}, LongVA-DPO~\citep{longva}, LongVU~\citep{longvu}, LLaVA-Video~\citep{llava-video} and Gemini-1.5-Pro~\citep{geminipro}, we sample 1 frame per second. For multi-modal agent, we evaluate VideoAgent~\citep{videoagent} and sample 1 frame every 2 seconds. All models (except for VideoAgent) are with an approximate total parameter size of 7B. 

\noindent
\textbf{Metrics and Benchmarks.}~~Since LongViTU answers are open-ended, for the evaluations on LongViTU test set, we designed a multi-level scoring system leveraging GPT-4’s near-human comprehension to evaluate answer-ground truth alignment. This approach rewards concise, accurate responses while penalizing hallucinations or irrelevant answers. Detailed criteria and prompts are in~\autoref{subsec:metrics}. Besides the LongViTU test set, we consider a collection of long-video understanding benchmarks. including ID benchmark EgoSchema~\citep{egoschema}, as well as OOD benchmarks VideoMME~\citep{videomme}, MLVU~\citep{mlvu}, and LVBench~\citep{lvbench}. Additionally, we also loop in canonical video understanding benchmarks (still OOD in terms of video sources), including MVBench~\citep{mvbench}, WorldQA~\citep{worldqa}, and OpenEQA~\citep{openeqa}.

\subsection{Quantitative Analysis on LongViTU Test Set}\label{subsec:exp_longvitu}

The quantitative results on LongViTU detailed in \autoref{tab:longvitu} reveal the following critical insights:

\noindent
\textbf{Challenges Posed by LongViTU.}~~Zero-shot evaluation of mainstream models on LongViTU reveals substantial challenges. Even after supervised fine-tuning (SFT), the best-performing model, LongVU, achieves only a GPT-4 score of 55.9, which is far below human performance at 81.0, underscoring a persistent gap. Additionally, proprietary models such as Gemini-1.5-Pro struggle with long-video comprehension, primarily because they must process full-length videos ranging from tens of minutes to hours. Moreover, these models exhibit a pronounced reliance on textual cues rather than spatiotemporal understanding, highlighting a fundamental limitation of current video-language models in handling long-video context and condensed reasoning challenges.

\subsection{Quantitative Analysis on Other Benchmarks}
\label{subsec:exp_others}

To evaluate the impact of SFT on LongViTU, we train several state-of-the-art video understanding models and assess their performance on both ID and OOD scenarios, as detailed in \autoref{tab:ood}. Key findings are outlined below.

\noindent
\textbf{Broad Effectiveness of SFT.}~~Fine-tuning on LongViTU leads to substantial performance improvements across nearly all ID, long video OOD, and non-long video OOD benchmarks for major open-source video understanding models. This indicates that the challenges posed by LongViTU (as outlined in~\autoref{tab:longvitu}) do not primarily arise from dataset bias, a claim further supported by human-level performance. Instead, LongViTU provides a wealth of diverse knowledge (long video context and condensed reasoning, discussed in~\autoref{sec:introduction}), equipping models with enhanced capabilities for video understanding across a broad range of scenarios.

\noindent
\textbf{Larger Gains on Longer Videos.}~~Fine-tuning with LongViTU demonstrates greater improvements for longer videos, as evident from performance trends across different subsets. While results declined on shorter videos, such as the VideoMME Short subset and OpenEQA, where the latter’s average duration (49s) is even shorter than VideoMME Short (83s), significant gains were observed in the Medium (563s) and Long (2386s) subsets. Notably, LLaVA-Video SFT improved by 1\% on VideoMME Long, whereas performance slightly decreased on Short (-0.1\%) and remained stable on Medium (+0.3\%). This pattern is consistent with broader trends in LongVU and LLaVA-Video, where the advantages of SFT become increasingly pronounced as video length increases. These results highlight LongViTU’s effectiveness in enhancing long video context comprehension, reinforcing the benefits of models optimized for extended temporal contexts despite minor trade-offs on shorter videos.

\noindent
\textbf{Task-Specific Enhancements.}~~Both LongVU and LLaVA-Video achieved significant gains in MLVU, primarily driven by improvements in the "Ego" task. These models also demonstrated strong capabilities in "Count" and "Order" tasks, though their performance in other areas remained relatively weaker. On LVBench, where most subcategories do not overlap, the most pronounced improvement was observed in "Temporal Grounding". A similar trend was evident in MVBench, where categories aligned with LongViTU’s predefined categories (notably in condensed reasoning tasks), such as "Episodic Reasoning" and "Unexpected Action", demonstrated significant gains, whereas improvements in other categories were more modest.

\subsection{Ablation Study on LongViTU}
\label{subsec:ablation}

Due to space constraints, additional details on the ablation study and qualitative evaluation are provided in~\autoref{sec:additional}. These analyses offer deeper insights into LongViTU’s adaptability to varying input lengths and its capacity to leverage extended training, highlighting its robustness and scalability.

\section{Conclusion}
\label{conclusion}

We've introduced LongViTU, a large-scale, automatically generated dataset for long-form video understanding, featuring long video context and condensed reasoning. Fine-tuning canonical VLMs on LongViTU yields substantial performance gains across both ID and diverse OOD long video understanding benchmarks, underscoring its effectiveness.


{
    \small
    \bibliographystyle{ieeenat_fullname}
    \bibliography{main}
}

\appendix
\clearpage
\setcounter{page}{1}
\maketitlesupplementary

\section{Limitation Statement and Future Works}
\label{sec:limitation}

\textbf{Limitation Statement.}~~Our automated pipeline exhibits adaptability across diverse scenarios, self-revision mechanism mitigates QA pairs with pronounced textual biases. However, as illustrated in~\autoref{fig:human_study}, human evaluation indicates that approximately 9\% of QA pairs still retain such biases. This limitation primarily arises from the reliance on LLM-generated multimodal data. Nevertheless, recent studies on vision-language models suggest that a controlled presence of such biases may enhance model robustness.

\noindent
\textbf{Future Works.}~~Future research will focus on enhancing data quality and exploring novel model architectures, \eg, learnable token compression and memory mechanism, \etc.


\section{Additional Results}
\label{sec:additional}

Due to space limitations, we defer more details about the ablation study and qualitative evaluation to this section.

\subsection{Ablation Study}
\label{subsec:ablation}

\begin{figure}[htbp]
    \centering
    \includegraphics[width=0.47\textwidth]{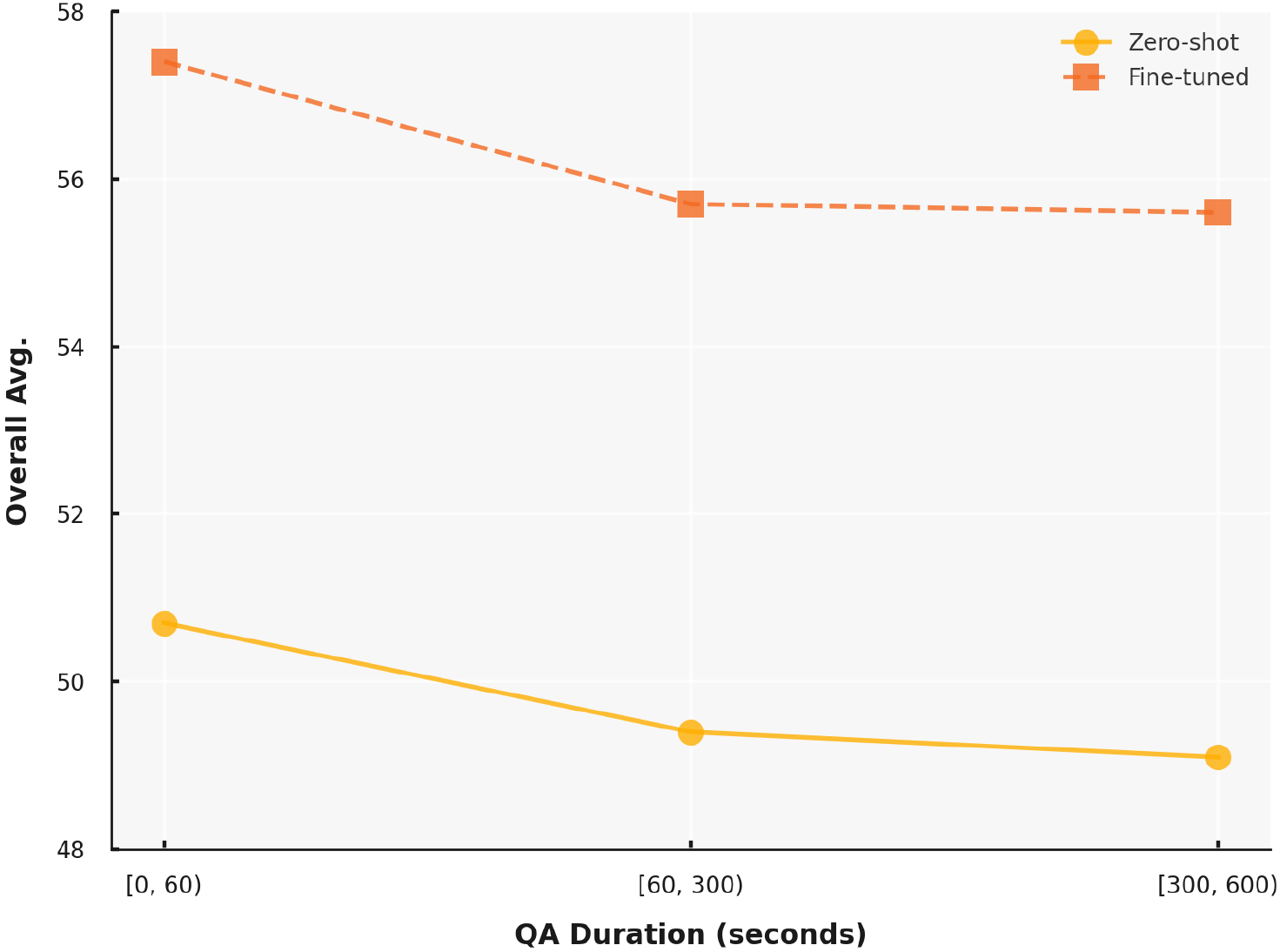}
    \caption{\textbf{Duration Ablation.}~~Regardless of whether in the zero-shot or fine-tuned setting on LongViTU, LongVU demonstrates a performance decline on longer subsets [300, 600) compared to shorter ones [60, 300) and [0, 60).}
    \label{fig:duration_ablation}
\end{figure}

\noindent
\textbf{Varying Durations.}~~\autoref{fig:duration_ablation} illustrates LongVU's performance on LongViTU with varying QA durations. As the QA duration (in seconds) increases, performance steadily declines, aligning with the trend observed in VideoMME across all baseline models (\autoref{tab:ood}). This downward trend persists across the Short, Medium, and Long subsets in both the zero-shot and SFT settings. This result is intuitive, as longer videos naturally introduce greater complexity and challenge compared to shorter ones.

\begin{figure}[htbp]
    \centering
    \includegraphics[width=0.47\textwidth]{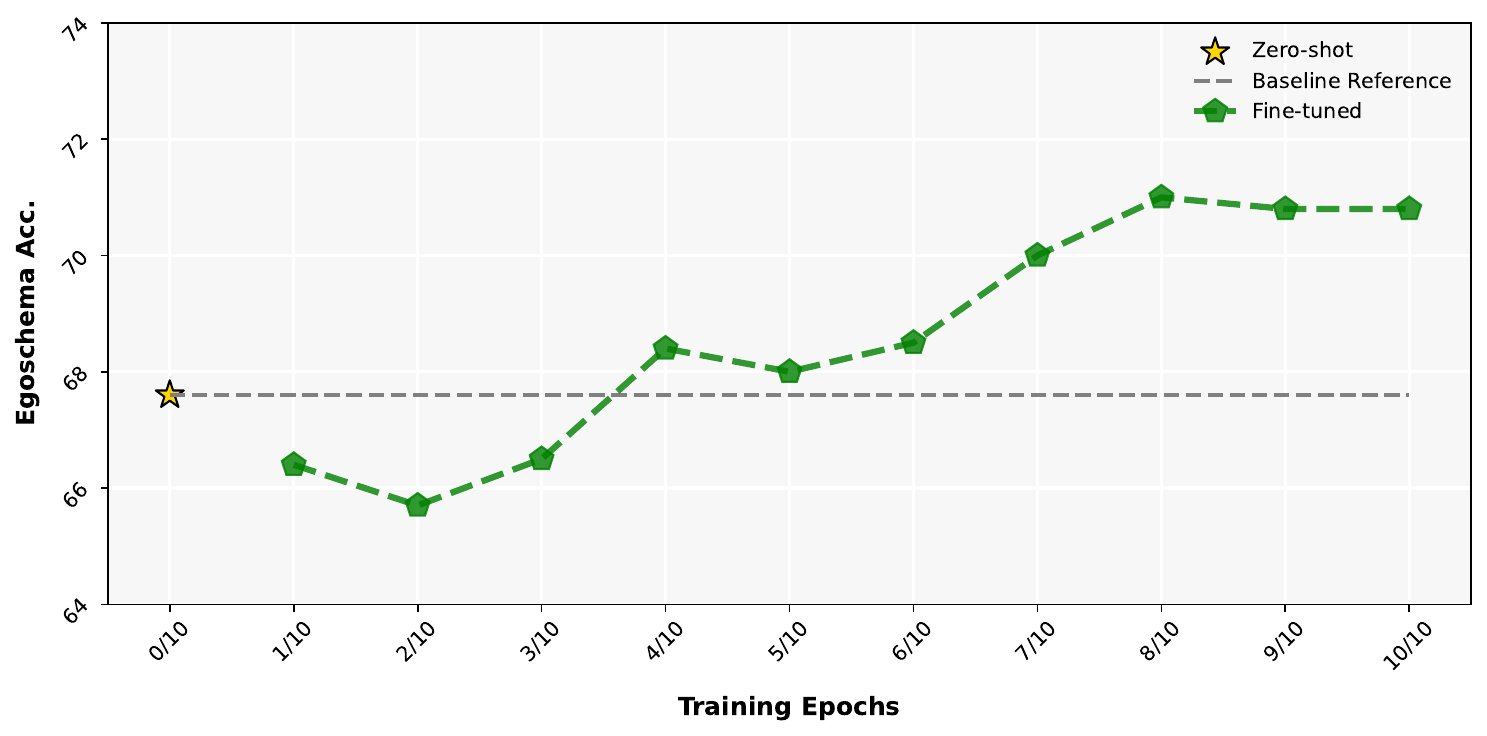}
    \caption{\textbf{Epoch Ablation.}~~Performance improvements during fine-tuning diminish as training epochs approach the full scale of LongViTU.}
    \label{fig:epoch_ablation}
\end{figure}

\noindent
\textbf{Training Epochs.}~~\autoref{fig:epoch_ablation} illustrates that LongVU's performance on Egoschema improves following SFT as the number of training epochs inherited from LongViTU increases. However, the performance gains gradually diminish as the epochs approach the full training scale of LongViTU, eventually stabilizing at its final performance.

\subsection{Qualitative Evaluation}
\label{subsec:qualitative}

\begin{figure*}[htbp]
    \centering
    \vskip 0.2in
    \begin{subfigure}[b]{1\textwidth}
        \centering
        \includegraphics[width=1\textwidth]{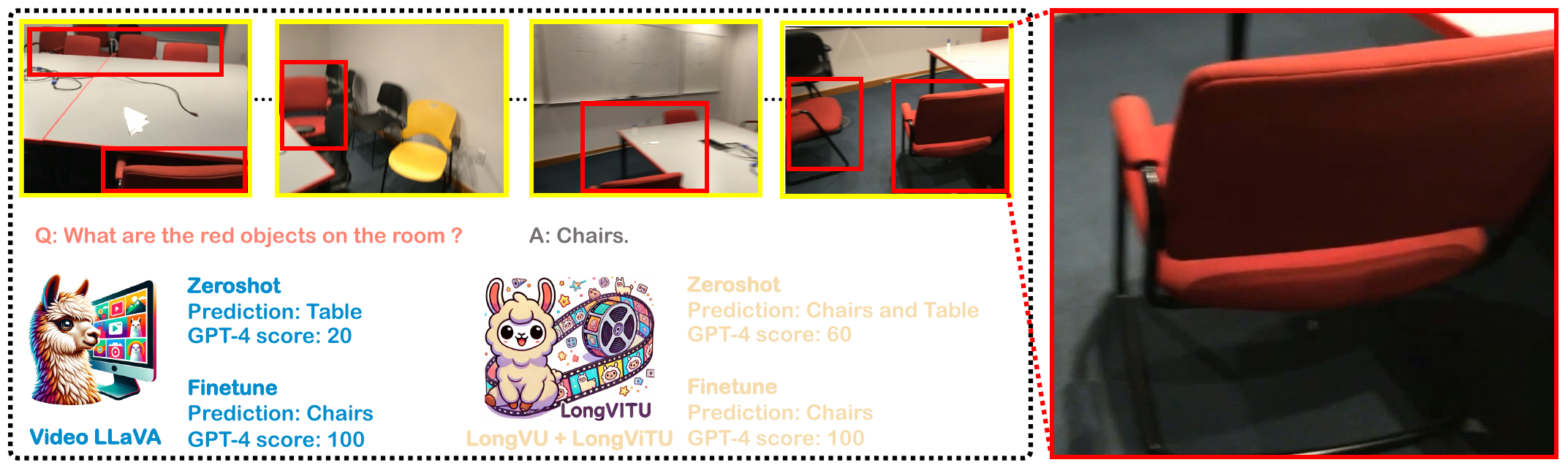}
        \caption{\textbf{Spatial Details.}~~The dense arrangement of multiple foreground objects in the scene led to incorrect zero-shot predictions by both Video-LLaVA and LongVU. Fine-tuning with LongViTU enabled the model to better capture fine-grained spatial features, resulting in correct answers.}
        \label{fig:qualitative_1}
    \end{subfigure}
    \vfill
    \vskip 0.2in
    \begin{subfigure}[b]{1\textwidth}
        \centering
        \includegraphics[width=1\textwidth]{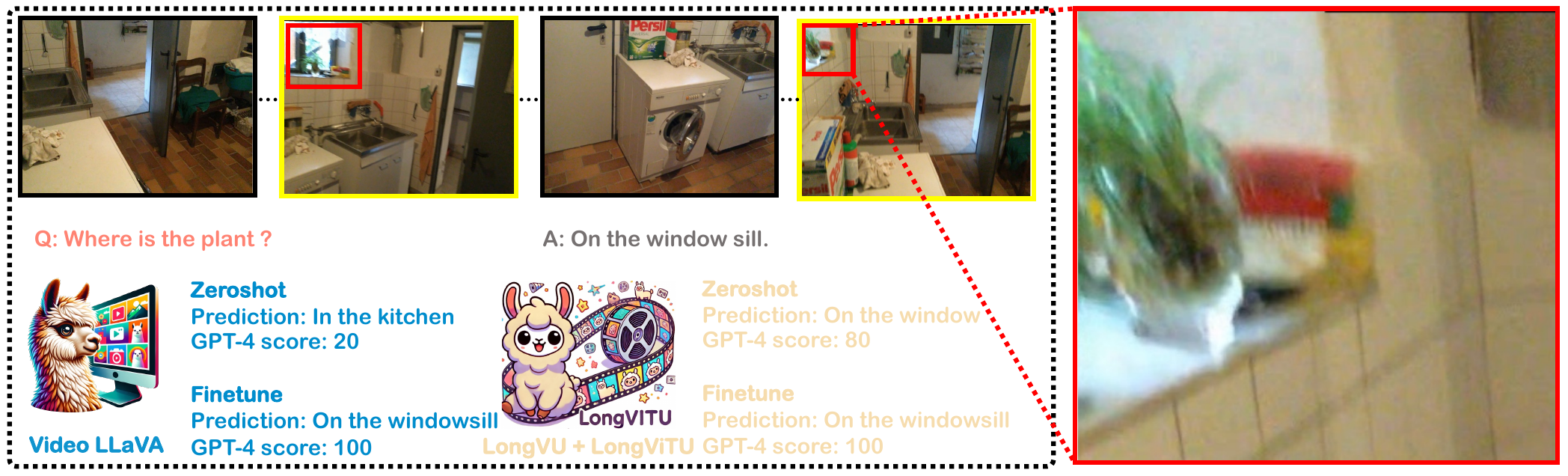}
        \caption{\textbf{Key Moments.}~~The fine-tuned models successfully identified the key moment, a brief appearance of \textit{"a plant on the windowsill"}, and provided a precise and concise response, achieving a perfect GPT-4 score of 100. In contrast, the Video-LLaVA zero-shot model failed to capture this brief key moment.}
        \label{fig:qualitative_2}
    \end{subfigure}
    \vfill
    \vskip 0.2in
    \begin{subfigure}[b]{1\textwidth}
        \centering
        \includegraphics[width=1\textwidth]{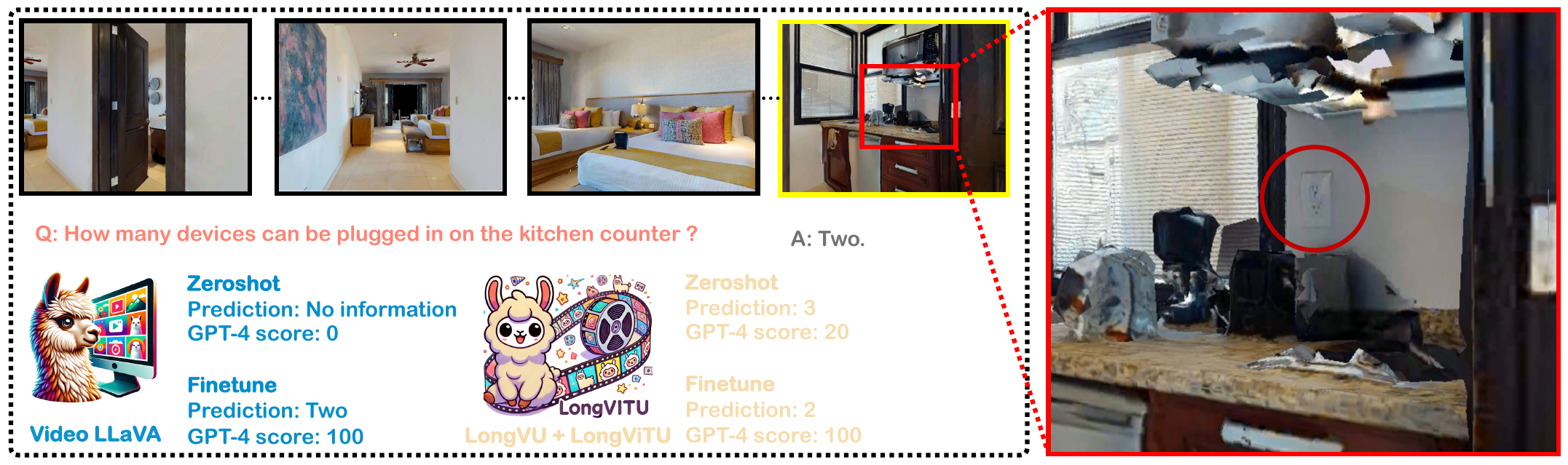}
        \caption{\textbf{Temporal Localization.}~~Both Video-LLaVA and LongVU accurately identified \textit{"two"} plug-in sockets in the kitchen at the end of a long video sequence following fine-tuning. However, in the zero-shot setting, both models struggled with this task. This observation underscores the challenge of extracting spatial details from extended video sequences and highlights the contribution of LongViTU in enhancing generalization for long-form temporal localization.}
        \label{fig:qualitative_3}
    \end{subfigure}
    \caption{\textbf{Qualitative Results.}~~The \colorbox{brightyellow!50}{yellow box} indicates the key frame that contains the answer, while the \colorbox{brightred!50}{red box} highlights the relevant objects. For better illustration, only concise key information is presented in the predictions.}
    \label{fig:qualitative}
\end{figure*}

To facilitate a more comprehensive qualitative analysis, we provide visualizations of various question-answering scenarios in~\autoref{fig:qualitative}. Due to space constraints, these figures have been deferred to this section.

\noindent
\textbf{Spatial Details.}~~As depicted in~\autoref{fig:qualitative_1}, the dense arrangement of multiple foreground objects in the scene led to incorrect zero-shot predictions by both Video-LLaVA and LongVU. Fine-tuning with LongViTU enabled the model to better capture fine-grained spatial features, resulting in correct answers.

\noindent
\textbf{Key Moments.}~~As illustrated in~\autoref{fig:qualitative_2}, the fine-tuned models successfully identified the key moment, a brief appearance of \textit{"a plant on the windowsill"}, and provided a precise and concise response, achieving a perfect GPT-4 score of 100. In contrast, the Video-LLaVA zero-shot model failed to capture this brief key moment.

\noindent
\textbf{Temporal Localization.}~~As shown in~\autoref{fig:qualitative_3}, both Video-LLaVA and LongVU accurately identified \textit{"two"} plug-in sockets in the kitchen at the end of a long video sequence following fine-tuning. However, in the zero-shot setting, both models struggled with this task. This observation underscores the challenge of extracting spatial details from extended video sequences and highlights the contribution of LongViTU in enhancing generalization for long-form temporal localization.

\section{More Details on Building LongViTU}
\label{sec:details}

\subsection{Hierarchical Video Tree Construction}

This subsection outlines the hierarchical video tree construction process, with the details provided in~\autoref{subsec:videotree}, and the algorithm elaborated in~\autoref{algo:hierarchical_video_tree_construction}. The corresponding prompts for each stage are described in the following.

\noindent
\textbf{Frame Level.}~~For dense captioning at the frame level, the \texttt{"internlm-xcomposer2-vl-7b-4bit"} model is employed with the following prompt:

\begin{tcolorbox}
\begin{minipage}{\linewidth}
<ImageHere>Identify each object in the image, describe their positions, and detail their appearance.
\end{minipage}
\end{tcolorbox}

\noindent
\textbf{Event Level.}~~We employ the \texttt{ChatCompletion} API of the \texttt{"gpt-4-turbo-2024-04-09"} model with the following prompt to refine event level descriptions:

\begin{tcolorbox}
\begin{minipage}{\linewidth}
Write a concise narrative in one sentence, including visual details from "Frames" that depict an "Event", do not use any unrelated information.\newline

"Event" describes an action in a video, with "C" representing me and other letters like 'X' and 'Y' standing for different people, transform these for a smoother narrative.\newline

"Frames" show detailed visuals and space details of objects in each moment during the "Event".\newline

Event:
\{event\}\newline
Frames:
\{frames\}\newline

Just return narrative that summarizes the episodic memory depicted in this video, only focuses on spatial details and temporal correlations.\newline

Narrative:
\end{minipage}
\end{tcolorbox}

\noindent
\textbf{Segment Level.}~~We utilize the \texttt{ChatCompletion} API of the \texttt{"gpt-4-turbo-2024-04-09"} model to generate segment level descriptions:

\begin{tcolorbox}
\begin{minipage}{\linewidth}
Integrate sequential event descriptions of video content into a very concise summary in one sentence, from my perspective for a smoother narrative. Each segment should capture a sequence of closely related actions, events, or scenes. Using "index" to represent the start and end of each segment, do not use any unrelated information.
\end{minipage}
\end{tcolorbox}

\begin{algorithm}[htbp]
\caption{Hierarchical Video Tree Construction}
\label{algo:hierarchical_video_tree_construction}
\begin{algorithmic}[1]
\REQUIRE Annotated events $Events$ in Ego4D
\ENSURE Hierarchical video tree $\mathcal{T}_{\text{video}}$
\STATE Initialize $\mathcal{T}_{\text{video}} = \emptyset$\; 
\STATE Sample frames $Frames$ at 1 fps across annotated events\; 
\FOR{each frame $f$ in $Frames$}
    \STATE $d_f = \text{InternLM-XComposer2}(f)$\; 
    \STATE $t_s^{f} = f.\text{timestamp}$; $t_e^{f} = f.\text{timestamp}$\; 
    \STATE $Frames[f] = \langle d_f, t_s^{f}, t_e^{f} \rangle$\;
\ENDFOR
\FOR{each event $e$ in $Events$}
    \STATE Collect frames $\{ \langle d_f, t_s^{f}, t_e^{f} \rangle \}$ within event $e$\;
    \STATE $d_e = \text{GPT-4}(\{ d_f \})$\; 
    \STATE $t_s^{e} = e.\text{start}$; $t_e^{e} = e.\text{end}$\; 
    \STATE $Events[e] = \langle d_e, t_s^{e}, t_e^{e}, \{ \langle d_f, t_s^{f}, t_e^{f} \rangle \} \rangle$\;
\ENDFOR
\STATE Group related events into segments $\{ Segments \}$ using GPT-4\; 
\FOR{each segment $s$ in $Segments$}
    \STATE $d_s = \text{GPT-4}(\{ d_e \text{ in } s \})$\; 
    \STATE $t_s^{s} = \min\{ t_s^{e} \text{ in } s \}$; $t_e^{s} = \max\{ t_e^{e} \text{ in } s \}$\; 
    \STATE $\mathcal{S}[s] = \langle d_s, t_s^{s}, t_e^{s}, \{ Events \text{ in } s \} \rangle$\;
\ENDFOR
\STATE $\mathcal{T}_{\text{video}} = \langle R, \{ \mathcal{S}[s] \} \rangle$\; 
\RETURN $\mathcal{T}_{\text{video}}$\;
\end{algorithmic}
\end{algorithm}

\begin{tcolorbox}
\begin{minipage}{\linewidth}
Step-by-step:\newline
1. Review event descriptions and group consecutive events that are closely related into a segment.\newline
2. For each group of events, write a brief summary.\newline

"index" represents order of event, "event" outlines this moment.\newline

Video Content:\newline
\{video content\}\newline

Return each segment in JSON format: {{"start": start index, "end": end index, "segment": brief description of video segment}}. Assemble all segments into a single Python list, ensuring output is neatly organized and strictly adheres to this JSON format.\newline

Segments:
\end{minipage}
\end{tcolorbox}

\subsection{Long-Form QA Generation}
\label{subsec:qa_generation}

\noindent
\textbf{QA Generation.}~~We adopt the \texttt{ChatCompletion} API of the \texttt{"gpt-4-turbo-2024-04-09"} model to generate QA pairs on the selected sub-tree:

\begin{tcolorbox}
\begin{minipage}{\linewidth}
Task:\newline
Construct episodic memory of video content through question-answer pairs that encapsulate spatial and temporal aspects within selected events.\newline

Step-by-Step Instructions:\newline
1. Selection of Events: Select either a single specific event or a series of interrelated events from the video content ('Memory Content'). For each selected event or sequence of events, generate question-answer pairs that reflect their spatial and temporal characteristics. Use "index" to designate the chronological order of these memory events.
2. Creation of Question-Answer Pairs: From the selected events, formulate questions that will be posed later in the video related to a single, specific event ('Ask Content'). These pairs should mimic a retrospective dialogue between me and an AI assistant, where I pose questions and the AI provides answers based on the video content. Reference events and segments to make dialogue more naturally narrative, avoiding direct references "index" or timestamps.\newline
3. Categorization of Questions: Categorize each question under a specific type such as: Object, Attribute, Location, Action, Function, Affordance, Comparison, Relationship, Causality, Motivation, Planning, Risk, or any other category you suggest.\newline

Output Format:\newline
Return question-answer pairs in JSON format: {{"memory": [list of memory events index], "ask": event index where question is posed, "type": question type, "question": question, "answer": answer}}. Assemble all pairs into a single Python list, ensuring the output is neatly organized and strictly adheres to this JSON format.\newline

Term Definitions of Video Content:\newline
- segment: a brief summary covering a sequence of related events.\newline
- events: multiple related events within a segment.\newline
- index: sequential position of an event within the overall video content.\newline
- event: spatial-temporal details associated with each moment in the video.\newline

Memory Content:\newline
\{memory content\}\newline
Ask Content:\newline
\{ask content\}\newline

Question-Answer pairs:
\end{minipage}
\end{tcolorbox}

\subsection{Self-Revision}
\label{subsec:selfrevision}

\noindent
\textbf{Self-Revision.}~~We utilize the \texttt{ChatCompletion} API of the \texttt{"gpt-4-turbo-2024-04-09"} model to perform self-revision:

\begin{tcolorbox}
\begin{minipage}{\linewidth}
Please review and correct the following question-answer pair about video content. Simplify the pair to directly represent the core information without redundant details, ensuring the question is natural and concise, and the answer is direct and clear.

Identify the correct type of the QA pair: Object, Attribute, Location, Action, Function, Affordance, Comparison, Relationship, Causality, Motivation, Planning, Risk, or Other. Do not add or fabricate content. Remove redundant event numbers and express the event directly.\newline

Original QA:\newline
\{original qa\}\newline

Return the Revised QA as a dict:\newline
\{"revised type": revised QA type, "revised question": revised question, "revised answer": revised answer\}\newline

Revised QA:
\end{minipage}
\end{tcolorbox}

\subsection{Evaluation Metrics}
\label{subsec:metrics}

\noindent
\textbf{Scoring Criteria.}~~We use the \texttt{ChatCompletion} API of the \texttt{"gpt-4-turbo-2024-04-09"} model to perform evaluation by designed scoring criteria:

\begin{tcolorbox}
\begin{minipage}{\linewidth}
As a scoring expert, your responsibility is to evaluate the accuracy of a model's response to a specific question about video content. You will be provided with the 'question' asked about the video, the 'answer' which is the correct answer based on the video, and the 'prediction' which is the model's response. Your task is to assess how accurately the model's 'prediction' answers the 'question' in relation to the 'answer'.\newline

Question:\newline
\{question\}\newline

Answer:\newline
\{answer\}\newline

Prediction:\newline
\{prediction\}\newline

Scoring Criteria:
\end{minipage}
\end{tcolorbox}

\begin{tcolorbox}
\begin{minipage}{\linewidth}
Level 1: The 'prediction' is unrelated to the 'question' or unintelligible, containing significant errors or irrelevant characters. Score: 0.\newline
Level 2: The 'prediction' is completely off-topic, not reflecting the factual content of the 'answer'. Score: 20.\newline
Level 3: The 'prediction' somewhat response the 'question' but includes errors or irrelevant details not found in the 'answer'. Score: 40.
Level 4: The 'prediction' generally response the 'question' but has some inaccuracies or irrelevant details compared to the 'answer'. Score: 60.\newline
Level 5: The 'prediction' accurately response the 'question' and is mostly consistent with the 'answer', with only minor discrepancies. Score: 80.\newline
Level 6: The 'prediction' perfectly response the 'question' and fully aligns with the facts provided in the 'answer'. Score: 100.\newline

Only provide the numerical score based on the criteria above without any additional commentary.\newline

Score:
\end{minipage}
\end{tcolorbox}

\subsection{Category Definitions}
\label{subsec:definitions}

To enhance Large Language Models (LLMs) in constructing episodic memory from video content, we employ a structured taxonomy for question generation. As described in~\autoref{subsec:stage2_qagen}, the LLM assigns each generated question to an appropriate category within this taxonomy, ensuring more condensed and targeted reasoning. Below, we provide the specific definitions and examples for each category.

\begin{itemize}
    \item \textbf{Object:}~~Questions identifying or referencing physical entities present in the video.  \\
    Question: What was the red object on the table?  \\
    Answer: A red coffee mug.
    
    \item \textbf{Attribute:}~~Questions focusing on descriptive properties (e.g., color, shape, texture, size) of objects or entities.  \\
    Question: What was unusual about the person's jacket?  \\
    Answer: It was bright yellow with reflective stripes.
    
    \item \textbf{Location:}~~Questions about the spatial placement of objects, characters, or events within the scene.  \\
    Question: Where was the book before it was picked up?  \\
    Answer: On the wooden shelf next to the lamp.
    
    \item \textbf{Action:}~~Questions regarding physical movements or activities performed by characters or objects.  \\
    Question: What did the man do after opening the door?  \\
    Answer: He walked inside and placed his bag on the table.
        
    \item \textbf{Transition:}~~Questions involving contrasts or similarities between objects, actions, or scenes.  \\
    Question: How did the first explosion differ from the second?  \\
    Answer: The first was small and controlled, while the second was much louder and caused debris to scatter.
    
    \item \textbf{Interaction:}~~Questions that explore connections between entities, including people, objects, or events.  \\
    Question: How was the old man related to the boy?  \\
    Answer: He was the boy’s grandfather.
    
    \item \textbf{Causality:}~~Questions about cause-and-effect relationships within events.  \\
    Question: Why did the lights go out suddenly?  \\
    Answer: Because a power surge occurred.
    
    \item \textbf{Motivation:}~~Questions exploring the intent or reasoning behind a character’s actions.  \\
    Question: Why did the thief hesitate before taking the wallet?  \\
    Answer: He noticed a security camera watching him.
    
    \item \textbf{Planning:}~~Questions related to future intentions or strategies within the video.  \\
    Question: What was the team’s plan to escape?  \\
    Answer: They decided to climb through the air ducts.
    
    \item \textbf{Risk:}~~Questions that examine potential danger, hazards, or consequences of an action.  \\
    Question: What was the risk of crossing the bridge?  \\
    Answer: It was old and unstable, with missing planks.

    \item \textbf{Function:}~~Questions addressing the purpose or intended use of an object or action.  \\
    Question: Why did the character press the button?  \\
    Answer: To call the elevator.
    
    \item \textbf{Affordance:}~~Questions about how an object can be used or interacted with, based on its design or properties.  \\
    Question: What did the woman use the chair for?  \\
    Answer: She stood on it to reach the top shelf.
\end{itemize}

\section{Implementations and Hyperparameters}
\label{sec:implementation}

The fine-tuning process for open-source models is governed by a set of key hyperparameters that influence their performance and generalization capabilities.~\autoref{tab:params} provides a structured overview of these critical training parameters for LLaMA-VID~\citep{llamavid}, Video-LLaVA~\citep{videollava}, LongVU~\citep{longvu}, and LLaVA-Video~\citep{llava-video}. Each model has been fine-tuned with distinct configurations to optimize its ability to process and understand video-based inputs effectively. Details regarding the pretraining strategies and implementation specifics of these models are anonymously available at \href{https://longvitu.github.io}{https://longvitu.github.io}.

\begin{table*}[t!]
    \centering
    \caption{\textbf{Fine-tuning Hyperparameters.}~~This table presents the essential hyperparameters employed during fine-tuning for the latest mainstream open-source models, all of which have 7B parameters for a fair comparison. Parameters that are not explicitly specified adhere to the default configurations of the respective model implementations.}
    \label{tab:params}

    \vspace{5mm}
    
    \begin{tabular}{ll|l}
      \toprule
      \multicolumn{2}{l|}{\textbf{Hyperparameter}} & \textbf{Value} \\

      \midrule
      \multicolumn{3}{l}{\textit{LLaMA-VID}} \\ 
      & Pretrained LLM & Vicuna-1.5-7B \\
      & Maximum sequence length & 2048 \\
      & Batch size & 1 \\
      & Gradient accumulation steps & 64 \\
      & Learning rate & $5e^{-6}$ \\
      & Weight decay & 0 \\
      & Warmup ratio & 0.03 \\
      & Learning rate scheduler & Cosine \\
      & Image tokens per sample & 2 \\
      & Video frame rate (FPS) & 1 \\

      \midrule
      \multicolumn{3}{l}{\textit{Video-LLaVA}} \\ 
      & Pretrained LLM & LLaMA-2-7B \\
      & Maximum sequence length & 2048 \\
      & Batch size & 32 \\
      & Gradient accumulation steps & 2 \\
      & Learning rate & $1e^{-5}$ \\
      & Weight decay & 0 \\
      & Warmup ratio & 0.03 \\
      & Learning rate scheduler & Cosine \\
      & Image tokens per sample & 256 \\
      & Video frames per sample & 8 \\

      \midrule
      \multicolumn{3}{l}{\textit{LongVU}} \\ 
      & Pretrained LLM & Qwen-2-7B \\
      & Maximum sequence length & 8192 \\
      & Batch size & 32 \\
      & Gradient accumulation steps & 1 \\
      & Learning rate & $5e^{-7}$ \\
      & Weight decay & 0 \\
      & Warmup ratio & 0.03 \\
      & Learning rate scheduler & Cosine \\
      & Image tokens per sample & 144 \\
      & Video frame rate (FPS) & 1 \\

      \midrule
      \multicolumn{3}{l}{\textit{LLaVA-Video}} \\ 
      & Pretrained LLM & Qwen-2-7B \\
      & Maximum sequence length & 32768 \\
      & Batch size & 1 \\
      & Gradient accumulation steps & 1 \\
      & Learning rate & $1e^{-7}$ \\
      & Vision tower learning rate & $2e^{-8}$ \\
      & Weight decay & 0 \\
      & Warmup ratio & 0.03 \\
      & Learning rate scheduler & Cosine \\
      & Image tokens per sample & 196 \\
      & Video frames per sample & 64 \\

      \bottomrule
    \end{tabular}
\end{table*}

\section{Related Work}
\label{sec:related}

\noindent
\textbf{Large Language Models.}~~Large language models (LLMs), including InstructGPT~\citep{instructgpt}, GPT-4~\citep{gpt4}, LLaMA~\citep{llama}, and LLaMA-2~\citep{llama2}, have demonstrated remarkable capabilities in text processing and large-scale dataset generation. These models convert multimodal inputs into structured textual representations, which can then be used to prompt GPT-4 for multimodal data synthesis.

\noindent
\textbf{Instruction Tuning.}~~LLaVA~\citep{llava} pioneered the use of foundational vision models to generate image captions and detect bounding boxes, which were subsequently processed by ChatGPT or GPT-4 to create image-based QA datasets. Building upon this, approaches such as Bongard-OpenWorld~\citep{bongard-openworld}, Video-LLaVA~\citep{videollava}, and VideoChat~\citep{videochat} extended these techniques to video-based QA. By sampling multiple frames from videos and applying LLaVA’s methodology, these methods generate structured video QA datasets using frame-wise descriptions, object categories, and attributes. However, the redundancy in frame-based descriptions, coupled with the input length constraints of LLMs, limits the number of frames that can be processed, thereby reducing dataset comprehensiveness.

\noindent
\textbf{Long-Context Language Models.}~~Despite advances in long-context LLMs~\citep{rope}, including GPT-4-turbo, ChatGLM~\citep{chatglm}, Baichuan2~\citep{baichuan2}, and InternLM2~\citep{internlm2}, these models exhibit significant performance degradation when processing extended and intricate texts. They struggle to manage the redundancy and structural disorganization inherent in detailed video descriptions, limiting their effectiveness in generating video QA. Unlike static images, video understanding requires modeling temporal dynamics, making event correlation essential. However, existing frame-based approaches fail to capture this aspect adequately, often resulting in shallow QA generation confined to frame-level analysis.

\noindent
\textbf{Long-Form Video Understanding.}~~Instruction tuning within the LLaVA paradigm~\citep{llava} has proven effective for multimodal tasks such as captioning~\citep{sharegpt4video} and visual question answering (VQA)~\citep{brown2020language, palm2, gemini, blip, instructblip, qwen2, flamingo, malmm, streamingvideo, videollamb}. While these methods perform well on images and short videos, they face significant challenges when applied to long-form videos~\citep{moviechat, videollava, videochatgpt, videollama, lvbench, longva}. A primary bottleneck arises from the large number of visual tokens produced by encoders—ranging from 576 to 2880 tokens per image in models like LLaVA-NeXT~\citep{llavanext}—which quickly exceed the context window limits of LLMs as the number of frames increases.
To address this, recent methods employ resampling techniques to reduce visual tokens before feeding them into LLMs~\citep{blip, llamavid, matryoshka, videollama2}. However, such reductions often degrade visual representations, leading to performance deterioration. Alternative strategies, including advanced token pruning and feature merging, offer promising solutions~\citep{chen2024image, llavaprumerge, chatunivi, streamingcaptioning}. Notably, recent models such as VideoLLM Online~\citep{videollmonline}, LongVILA~\citep{longvila}, LLaVA-OV~\citep{llava-ov}, LongVA~\citep{longva}, and LongVU~\citep{longvu} represent significant advancements in addressing these limitations and enhancing long-form video understanding.


\section{More LongViTU Examples}
\label{sec:examples}

The following presents more examples of the detailed categorization of LongViTU.

\begin{figure*}[htbp]
    \centering
    \vskip 0.2in
    \begin{subfigure}[b]{1\textwidth}
        \centering
        \includegraphics[width=1\textwidth]{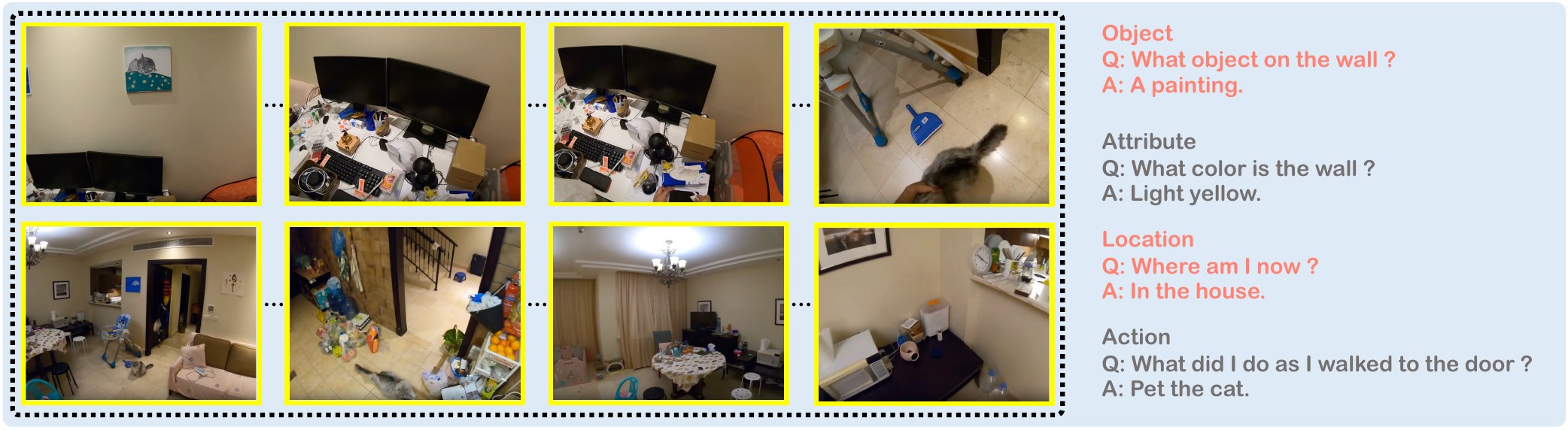}
        \caption{\textbf{Spatiotemporal Understanding.}~~This subfigure presents examples for four subcategories within Spatialtemporal Understanding, including Object, Attribute, Location, Action.}
        \label{fig:example_1}
    \end{subfigure}
    \vfill
    \vskip 0.2in
    \begin{subfigure}[b]{1\textwidth}
        \centering
        \includegraphics[width=1\textwidth]{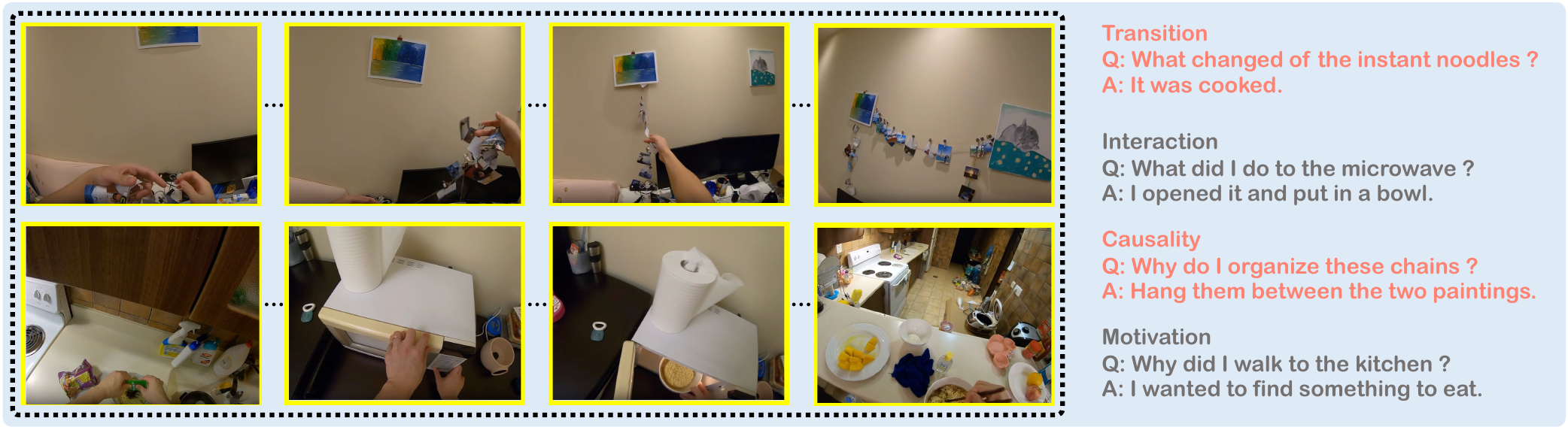}
        \caption{\textbf{Episodic Reasoning.}~~This subfigure presents examples for four subcategories within Episodic Reasoning, including Transition, Interaction, Causality, Motivation.}
        \label{fig:example_2}
    \end{subfigure}
    \vfill
    \vskip 0.2in
    \begin{subfigure}[b]{1\textwidth}
        \centering
        \includegraphics[width=1\textwidth]{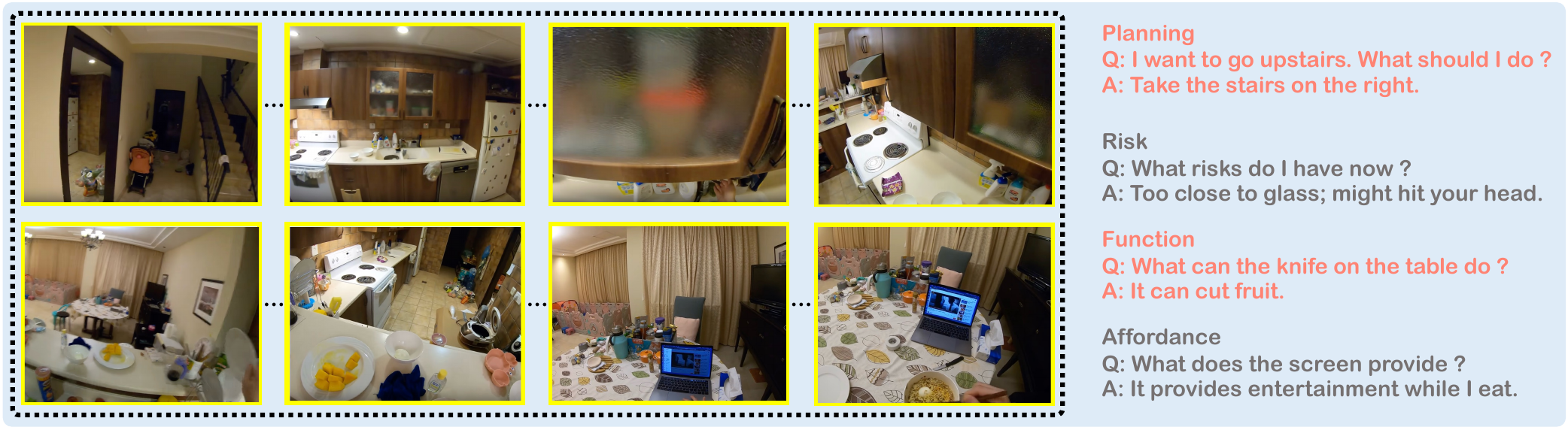}
        \caption{\textbf{Commonsense Inference.}This subfigure presents examples for four subcategories within Commonsense Inference, including Planning, Risk, Function, Affordance.}
        \label{fig:example_3}
    \end{subfigure}
    \caption{\textbf{LongViTU Examples.}~~For better illustration, only concise key information is presented in the predictions.}
    \label{fig:examples}
\end{figure*}

\end{document}

%% file: config.tex
\definecolor{Gray}{gray}{0.9}
\definecolor{mygreen}{rgb}{0.0, 0.5, 0.0}
\definecolor{myred}{rgb}{0.8, 0.25, 0.33}
\definecolor{myblue}{rgb}{0.19, 0.55, 0.91}
\definecolor{uclablue}{rgb}{0.15, 0.45, 0.68}
\definecolor{ucladblue}{rgb}{0.0, 0.33, 0.53}
\definecolor{ucladdblue}{rgb}{0.0, 0.23, 0.36}
\definecolor{uclagold}{rgb}{1.0, 0.82, 0.0}
\definecolor{ucladgold}{rgb}{1.0, 0.78, 0.17}
\definecolor{ucladdgold}{rgb}{1.0, 0.72, 0.11}
\definecolor{boxgreen}{rgb}{0.02, 0.66, 0.02}
\definecolor{boxred}{rgb}{0.66, 0.1, 0.1}
\definecolor{boxblue}{rgb}{0.01, 0.01, 0.73}

\usepackage{etoc}
\usepackage{lipsum}
\usepackage{wrapfig}
\usepackage{multicol}
\usepackage{mdframed}

\usepackage{mathtools}

\usepackage[utf8]{inputenc}   
\usepackage[T1]{fontenc}      
\usepackage{url}              
\usepackage{amsfonts}         
\usepackage{nicefrac}         
\usepackage{microtype}
\usepackage{graphicx}
\usepackage{amsmath,amssymb,mathbbol}
\usepackage{algorithmic}
\usepackage[linesnumbered,ruled,vlined]{algorithm2e}
\usepackage{acronym}
\usepackage{enumitem}
\usepackage{balance}
\usepackage{xspace}
\usepackage{setspace}
\usepackage[skip=3pt,font=small]{subcaption}
\usepackage[skip=3pt,font=small]{caption}
\usepackage[dvipsnames]{xcolor}
\usepackage{booktabs,tabularx,colortbl,multirow,array,makecell}
\usepackage{pifont}
\usepackage{xr-hyper}
\usepackage{tcolorbox}
\tcbuselibrary{breakable}

\usepackage{siunitx}   
\usepackage{soul}   
\usepackage{listings}
\definecolor{codegreen}{rgb}{0,0.6,0}
\definecolor{codegray}{rgb}{0.5,0.5,0.5}
\definecolor{codepurple}{rgb}{0.58,0,0.82}
\definecolor{backcolour}{rgb}{0.95,0.95,0.92}

\lstdefinestyle{mystyle}{
    backgroundcolor=\color{backcolour},   
    commentstyle=\color{codegreen},
    keywordstyle=\color{magenta},
    numberstyle=\tiny\color{codegray},
    stringstyle=\color{codepurple},
    basicstyle=\ttfamily\footnotesize,
    breakatwhitespace=false,         
    breaklines=true,                 
    captionpos=b,                    
    keepspaces=true,                 
    numbers=left,                    
    numbersep=5pt,                  
    showspaces=false,                
    showstringspaces=false,
    showtabs=false,                  
    tabsize=2
}

\makeatletter
\renewcommand{\paragraph}{%
  \@startsection{paragraph}{4}%
  {\z@}{0ex \@plus 0ex \@minus 0ex}{-1em}%
  {\hskip\parindent\normalfont\normalsize\bfseries}%
}
\makeatother

\definecolor{gblue}{HTML}{4285F4}
\definecolor{gred}{HTML}{DB4437}
\definecolor{ggreen}{HTML}{0F9D58}

\definecolor{mygray}{gray}{.92}

\newcommand{\cmark}{\ding{51}}
\newcommand{\xmark}{\ding{55}}

\acrodef{llm}[LLMs]{large language models}
\acrodef{vla}[VLA]{vision-language-action}
\acrodef{vl}[VL]{vision-language}
\acrodef{ocot}[O-CoT]{Object-centric Chain-of-Thought}
\acrodef{cot}[CoT]{Chain-of-Thought}
\acrodef{lvlm}[LVLM]{large vision-language models}